%% file: main.tex
\documentclass[10pt,twocolumn,letterpaper]{article}

 \usepackage{cvpr}       % To produce the REVIEW version
\usepackage{fvextra}
\usepackage{makecell}
\input{preamble}
\usepackage[accsupp]{axessibility}
\definecolor{cvprblue}{rgb}{0.21,0.49,0.74}
\usepackage[pagebackref,breaklinks,colorlinks,allcolors=cvprblue]{hyperref}

\def\paperID{27} % *** Enter the Paper ID here
\def\confName{CVPR}
\def\confYear{2026}

\title{Unsafe2Safe: Controllable Image Anonymization for Downstream Utility}

\author{\textbf{Minh Dinh} \quad \textbf{SouYoung Jin}\\
Dartmouth College\\
{\tt\small \{Minh.T.Dinh.GR, SouYoung.Jin\}@dartmouth.edu}
}
\begin{document}
\pagenumbering{arabic}
\setcounter{page}{1}
\maketitle
\input{sec/0_abstract_v1}    
\input{sec/1_intro_v2}

\input{sec/2_relatedwork_v1}
\input{sec/3_methods}
\input{sec/4_results}

\input{sec/5_conclusion}

\clearpage

\input{sec/6_acknowledgments}

{
    \small
    % \balance
    \bibliographystyle{ieeenat}
    \bibliography{main}
}

\input{sec/X_suppl}


\end{document}

%% file: preamble.tex
\usepackage{array}
\usepackage{tabularx}
\usepackage{tikz}
\usetikzlibrary{positioning}
\usepackage{ragged2e}

\newcommand{\sysname}{{Unsafe2Safe}\xspace}

\usepackage{multirow}
\usepackage{xcolor}
\usepackage{colortbl}

\definecolor{bestcell}{RGB}{255,249,196}   % light yellow
\definecolor{secondcell}{RGB}{220,240,255} % light blue

\definecolor{yesgreen}{RGB}{0,120,0}
\definecolor{partialorange}{RGB}{185,120,0}
\usepackage{pifont} 
\newcommand{\prompt}[1]{%
\colorbox{gray!15}{\texttt{#1}}%
}

\newcommand{\changelogSJ}[1]{\textcolor{black}{#1}}
\newcommand{\changelogSJcamera}[1]{\textcolor{black}{#1}}

%% file: sec/0_abstract_v1.tex
\begin{abstract}
Large-scale image datasets frequently contain identifiable or sensitive content, raising privacy risks when training models that may memorize and leak such information. We present \textbf{Unsafe2Safe}\footnote{\hyperlink{https://see-ai-lab.github.io/unsafe2safe/}{https://see-ai-lab.github.io/unsafe2safe/}}, a fully automated pipeline that detects privacy-prone images and rewrites only their sensitive regions using multimodally guided diffusion editing. Unsafe2Safe operates in two stages. \textbf{Stage~1} uses a vision--language model to (i) inspect images for privacy risks, (ii) generate paired \emph{private} and \emph{public} captions that respectively include and omit sensitive attributes, and (iii) prompt a large language model to produce structured, identity-neutral edit instructions conditioned on the public caption. \textbf{Stage~2} employs instruction-driven diffusion editors to apply these dual textual prompts, producing privacy-safe images that preserve global structure and task-relevant semantics while neutralizing private content. To measure anonymization quality, we introduce a unified evaluation suite covering \emph{Quality}, \emph{Cheating}, \emph{Privacy}, and \emph{Utility} dimensions. Across MS-COCO, Caltech101 and MIT Indoor67, Unsafe2Safe reduces face similarity, text similarity, and demographic predictability by large margins, while maintaining downstream model accuracy comparable to training on raw data. Fine-tuning diffusion editors on our automatically generated triplets (private caption, public caption, edit instruction) further improves both privacy protection and semantic fidelity. Unsafe2Safe provides a scalable, principled solution for constructing large, privacy-safe datasets without sacrificing visual consistency or downstream utility.
\end{abstract}

%% file: sec/1_intro_v2.tex
\section{Introduction}
\label{sec:intro}

\input{figures/teaser_figure}

Large-scale image datasets underpin modern computer vision, yet they contain faces, ID badges, personal documents, or other signals that expose individuals to privacy risks. These risks are amplified by the tendency of deep networks to memorize training data~\cite{staab2023beyond,wei2024evaluating} and thus are vulnerable to sensitive information extraction through model inversion~\cite{ma2025inversion,shan2025geminio} or membership attack~\cite{fu2024membership,samira2025variance}. As datasets grow larger and more heterogeneous, exhaustive manual auditing becomes infeasible, forming a barrier to sharing data and training models responsibly at scale.

Existing anonymization methods offer only partial protection. Preceeding redaction, e.g. blurring, masking, and inpainting, there is often a detector that often overlooks privacy cues outside predefined regions and introduces artifacts that impair downstream utility. The core challenge is that anonymization is not merely a removal problem: sensitive content must be rewritten so that identity cues are suppressed without destroying the semantic structure needed for learning. An effective solution must therefore combine strong privacy protection with semantic fidelity and broad applicability in open-domain images, without relying on expensive manual annotation. Anonymization thus requires fine-grained control over \emph{which} pixels to edit and \emph{how} to change them without compromising non-private semantics.

In this work, we introduce \textbf{\sysname{} (U2S)}, a scalable anonymization framework that uses multimodal reasoning and text-guided diffusion to produce privacy-safe yet utility-preserving versions of image datasets. Our key idea is to combine two complementary textual signals that jointly capture the semantics of the original scene while specifying how sensitive content should be altered. \changelogSJ{The overall process proceeds in two steps: first generating privacy-aware textual guidance, and then applying controlled image editing conditioned on this guidance.}

\changelogSJ{Firstly, a vision--language model (VLM) inspects each image using predefined privacy criteria and flags those containing sensitive content. For unsafe images, it produces two captions: a \emph{private caption} describing the full scene and a \emph{public caption} that removes identity-specific details while preserving non-sensitive semantics. As the public caption specifies the desired safe outcome but not the transformation itself, a large language model (LLM) further generates a structured \emph{edit instruction} that proposes identity-neutral substitutions and defines the required modifications. Together, the public caption and edit instruction determine what should be preserved and what should be altered.}

\changelogSJ{Conditioned on these textual signals, a text-guided diffusion editor performs targeted anonymization while maintaining visual coherence. The public caption acts as a semantic anchor that preserves global layout and object relationships, whereas the edit instruction guides geometry- and style-consistent rewriting of sensitive attributes. Unlike detector-based pipelines, \sysname{} does not require segmentation masks, attribute labels, or predefined privacy taxonomies. Instead, it leverages modern diffusion editors such as InstructPix2Pix\cite{brooks2023instructpix2pix} and FlowEdit\cite{kulikov2025flowedit}, which support multi-prompt conditioning to modify only the regions implied by the instructions. This design yields privacy-neutral reconstructions, removing identity cues and other sensitive or unsafe attributes while retaining structural fidelity and avoiding the semantic drift often observed in full image regeneration. Figure~\ref{fig:teaser} illustrates how \sysname{} selectively anonymizes sensitive regions while preserving recognizability and downstream task relevance.}

Through extensive experiments on Caltech-101~\cite{caltech101}, MIT Indoor-67~\cite{MITindoor67}, \changelogSJ{MS-COCO~\cite{lin2014microsoft} and OK-VQA~\cite{okvqa}}, we show that \sysname{} achieves strong anonymization while maintaining or even improving downstream task performance. Models trained on Unsafe2Safe data match the accuracy of models trained on raw images, and in some cases exceed them due to removal of spurious correlations. \sysname{} also reduces face similarity and text similarity, increases demographic diversity, and avoids semantic drift. Compared to baseline anonymization techniques, our approach leverages multimodal reasoning, dual-caption conditioning, and targeted diffusion editing, which together provide finer control, higher fidelity, and better scalability.

\vspace{0.1cm}
Our main contributions are summarized as follows:
\begin{enumerate}[label=(\arabic*)]

    \item We introduce \textbf{Unsafe2Safe}, a controllable privacy-preserving diffusion 
    framework that combines VLM-guided privacy inspection, public caption generation, 
    and LLM-derived edit instructions to remove identity-sensitive content while 
    preserving task-relevant semantics and spatial layout.

    \item We develop a \textbf{scalable anonymization pipeline and dataset construction process}
    that enables safe proxy generation across diverse domains.  
    We release privacy-safe generated datasets along with tools for anonymizing additional datasets, supporting reproducibility and broad community adoption.
%\footnote{All datasets and code will be released.}

    \item We propose a \textbf{unified evaluation framework} for anonymization, introducing
    four metric groups that jointly quantify image quality, leakage pathways, demographic
    diversity, and downstream utility.

    \item Through extensive experiments on \changelogSJ{Caltech-101, MIT Indoor-67, MS-COCO, and OK-VQA}, we show that \sysname{} \textbf{preserves or improves downstream accuracy} while 
    \textbf{substantially reducing face similarity, text similarity, and demographic 
    predictability}, outperforming existing anonymization baselines.

\end{enumerate}

%% file: figures/teaser_figure.tex
\begin{figure}[]
\centering
\vspace{-10pt}
%\fbox{\rule{0pt}{2in} \rule{.95\linewidth}{0pt}}
\resizebox{0.9\columnwidth}{!}{\includegraphics[trim=0pt 0pt 50pt 0pt, clip,keepaspectratio]{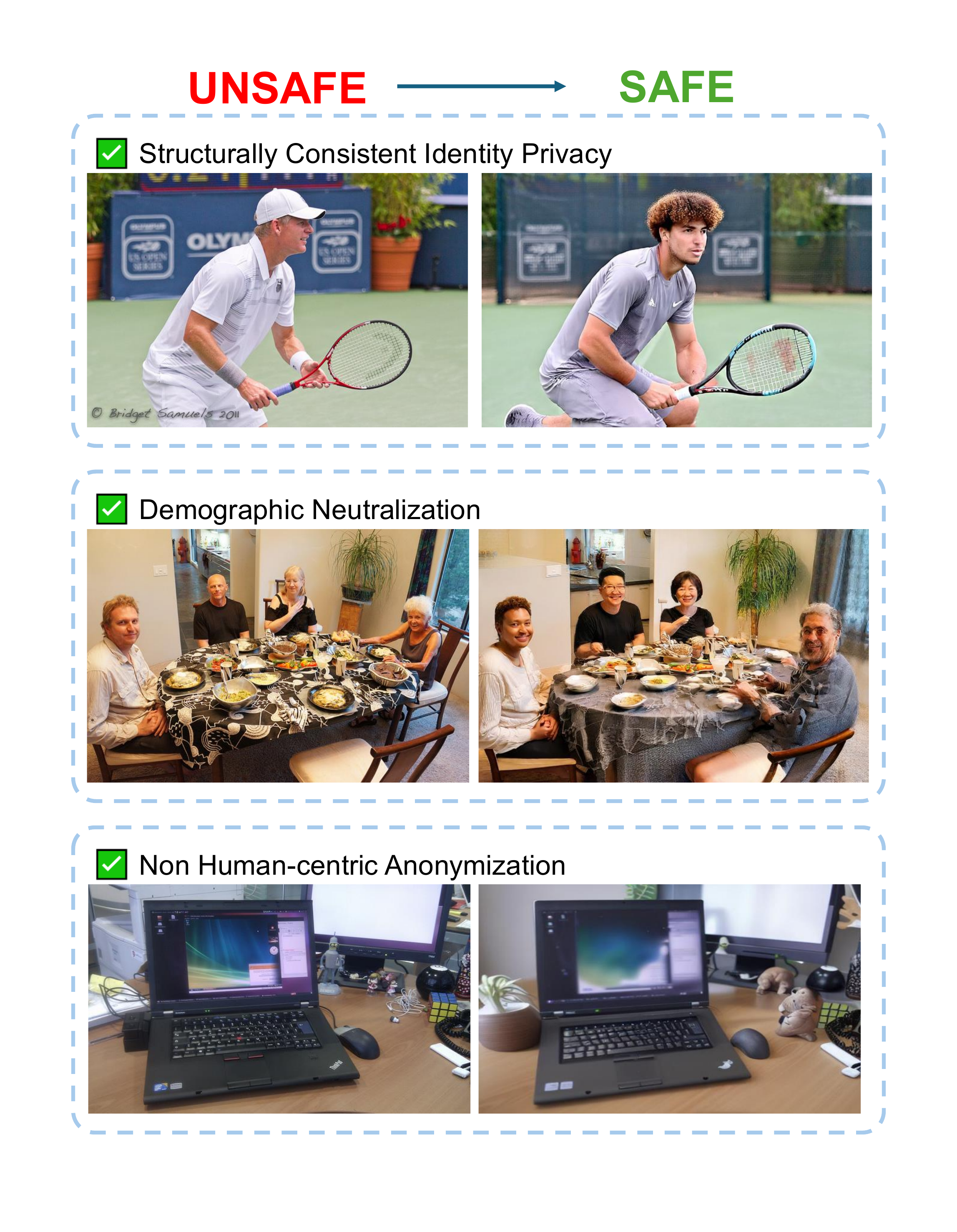}}
\vspace{-0.25in}
\caption{
Examples from \sysname{} (U2S).  
For each case, the model converts an \emph{unsafe} image into a privacy-preserving \emph{safe} version.  
\changelogSJ{Examples demonstrate key capabilities that may appear simultaneously:} (1) structure-preserving full body anonymization, (2) demographic neutralization (race entropy $\uparrow$), and (3) obfuscation of non-human confidential details.
}\vspace{-0.2in}
\label{fig:teaser}
\end{figure}

%% file: sec/2_relatedwork_v1.tex
\section{Related Work}

\subsection{Privacy-Preserving Data Generation}

Our work follows a data-centric approach to privacy: instead of protecting models after training, we anonymize images beforehand so that sensitive information never enters model weights. Most anonymization pipelines rely on face or person detectors~\cite{hukkelaas2019deepprivacy, hukkelaas2023deepprivacy2, kung2025face, hadera2025blanket} and then apply blurring, masking, pixelation~\cite{grauman2022ego4d, murrugarrabeyond}, or generative inpainting~\cite{patwari2024rendering, hukkelaas2023deepprivacy2}. Others detect risks via embedding similarity~\cite{ilic2024selective}. Yet, their reliance on the closed-set detector limits their applicability to open-domain imagery where privacy-encoding signals extend far beyond faces.

VLMs and LLMs \cite{yang2025qwen3, chen2024expanding, grattafiori2024llama, achiam2023gpt} offer a promising alternative for 
zero-shot privacy inspection and natural-language reasoning about sensitive content~\cite{tomekce2024private, murrugarrabeyond}. However, existing pipelines use such models only for detection, without leveraging their semantic understanding to guide how sensitive content should be modified.  \sysname fills this gap by coupling VLM-based privacy inspection with LLM-generated edit instructions and executing the transformation using text-guided diffusion editing, fostering structured, context-aware anonymization in open-set categories.

\subsection{Privacy Evaluation Frameworks}

Prior work measures privacy via re-identification attacks~\cite{kung2025face, egin2025now} or identity/attribute classification degradation~\cite{Dave_2022_CVPR, fioresi2023tedspad}, but treats privacy and utility as separate objectives. Datasets that annotate both~\cite{bega2020IPNhand, 1334462} are limited in scale, often restricted to binary human-centric attributes, and predominantly video-based, where utility depends on temporal cues rather than fine-grained spatial semantics.

More comprehensive benchmarks~\cite{abdulaziz2025evaluation, murrugarrabeyond} combine privacy and utility considerations but rely on human-annotated privacy labels and automatically generated utility labels, an inversion of real needs: privacy can often be inferred from general knowledge, but utility annotations require domain expertise and are costly. \textbf{Our evaluation approach reverses this imbalance.} We use VLMs for scalable, zero-shot privacy judgments and use original downstream datasets to measure utility. The resulting framework jointly assesses: (1) semantic fidelity, (2) residual privacy leakage, (3) fairness via demographic diversity, and (4) downstream performance when training on anonymized data—aligning evaluation with real-world deployment constraints.

\subsection{Controllable Image Editing}

\textbf{Anonymization with Diffusion.}
Recent diffusion-based anonymization focuses on identity-neutral face synthesis~\cite{hukkelaas2023deepprivacy2, kung2025face}, sometimes incorporating structural controls such as ControlNet~\cite{zhang2023adding} or ReferenceNet~\cite{hu2023animateanyone}. While these improve realism, they still rely heavily on masks, detectors, or handcrafted attributes, limiting their ability to handle arbitrary privacy cues. Mask-based redaction also introduces artifact boundaries and harms downstream utility.

\noindent
\textbf{Diffusion Editing.}
Advances in diffusion editing~\cite{brooks2023instructpix2pix, zhang2023magicbrush, zhao2024ultraeditinstructionbasedfinegrainedimage, Liu_2024_CVPR} allow text-guided modifications via hybrid conditioning or attention manipulation. However, most operate on UNet backbones with limited long-range modeling. Diffusion Transformers (DiTs)~\cite{esser2024scaling, flux2024, labs2025flux1kontextflowmatching} improve fidelity and flexibility, inspiring both training-based~\cite{tan2025ominicontrol, Zhang_2025_ICCV, tan2025ominicontrol2} and training-free editors~\cite{kulikov2025flowedit, yan2025eeditrethinkingspatial}. FlowEdit~\cite{kulikov2025flowedit}, in particular, uses ODE-based paths for structure-preserving transformations. Despite this progress, existing editors lack a mechanism for determining what is sensitive or how to modify it while preserving non-private semantics.
\sysname is the first to integrate VLM-driven privacy reasoning, LLM edit-generation, and modern diffusion editing into a unified, scalable anonymization pipeline for open-domain images.

%% file: sec/3_methods.tex
\section{\sysname: A Privacy-Preserving Pipeline for Visual Data}

\label{sec:method}

% We introduce \sysname, an automatic pipeline that detects unsafe images and anonymizes them into \textit{safe} versions while preserving utility for downstream tasks.
% \sysname{} identifies \textit{unsafe} images and transforms them into \textit{safe} alternatives using modern diffusion-based generative models.
We introduce \sysname, an automatic pipeline that detects \textit{unsafe} images and transforms them into \textit{safe} versions using modern diffusion-based generative models, while preserving utility for downstream tasks.

Recent diffusion models, such as Stable Diffusion and its variants~\cite{kulikov2025flowedit,brooks2023instructpix2pix}, have demonstrated strong capability in translating visual content across a wide range of modalities when guided by textual conditions. However, these models require an instruction $k$ that specifies how the model should re-generate an image $I$. To the best of our knowledge, no dataset currently provides such edit instructions. We therefore propose a novel approach that automatically generates captions describing the source image $I$ and the corresponding edit instructions needed for safe regeneration.

Building on these generative capabilities, our objective is to develop a diffusion-based framework that transforms privacy-prone images into privacy-safe counterparts while preserving both structural and semantic fidelity.

\input{figures/pipeline}

\sysname{} processes an image dataset and anonymizes unsafe regions by replacing or modifying private content according to a \textit{safe} description, while leaving non-private and task-relevant areas unchanged. To achieve this, \sysname{} operates in two stages, as illustrated in Figure~\ref{fig:pipeline}. \textbf{Stage~1} contains three components: \textit{image privacy inspection}, where a VLM identifies privacy-sensitive images; \textit{image captioning}, where the same VLM produces both private and public captions describing the image with and without sensitive attributes; and \textit{edit instruction generation}, where an LLM creates neutral, identity-free modification prompts based on the public captions. \textbf{Stage~2} then applies an image editor to generate the final safe images.

\subsection{Stage 1: Inspection}
\label{sec:datacurate}

\paragraph{Image Privacy Inspection.}
\label{sec:privacy_inspection}
We provide a VLM agent with a predefined set of privacy criteria and ask it to inspect each image to determine whether any criterion is present. If so, the image is marked as unsafe, and the system returns \verb|PRIVACY_FLAG=TRUE|. To minimize the risk of missing private content, we deliberately allow a higher Type~I error rate (false positives). The criteria used are derived from the private attribute set in VISPR~\cite{orekondy17iccv}, which consolidates attributes based on both regulatory guidelines and widely accepted norms in cyberspace.
\vspace{-10pt}
\paragraph{Image Captioning.}
\label{sec:captioning}
Using the \verb|PRIVACY_FLAG| from the inspection step, we separate the images into those containing privacy risks and those deemed safe. For each private image, the VLM generates two captions:
\begin{enumerate}
    \item \textbf{Private caption} ($c^{priv}$): fully describes the scene, including private attributes.
    \item \textbf{Public caption} ($c^{pub}$): describes the same scene while omitting all private details.
\end{enumerate}
The public caption $c^{pub}$ serves as a modality-aligned, privacy-preserving representation of the image and is later used as the base condition for instruction generation and safe image synthesis.
\vspace{-10pt}

\paragraph{Edit Instruction Generation.}
\label{sec:edit_instruction}
Public captions $c^{pub}$ describe the source image but do not provide any information about how to edit the image to produce a safe alternative. They realistically capture the scene while omitting all private details, making them an ideal canvas for controlled attribute insertion. To generate meaningful editing guidance, we leverage an LLM to enrich the public captions with plausible pseudo-private details. 
The LLM is prompted to generate plausible pseudo-private attributes that could apply to the scene, without relying on a fixed predefined list, and returns an edit instruction $c^{edit}$ describing how the image should be modified. We  employ an LLM to merge the two into a compact, instruction-style prompt that preserves both semantic grounding and 
edit-specific guidance, referring it as $LLM$ text prior.

    % \item \textbf{Caption integration}: The LLM merges $c^{edit}$ with the public caption $c^{pub}$ to form a unified caption $c^{pub\_detailed}$ that preserves the public description while incorporating the proposed edits.

% The resulting instruction $c^{edit}$ is later merged with the public caption $c^{pub}$ to form a unified caption $c^{pub\_detailed}$, which is used as the conditioning text for the diffusion model in Stage~2.

\subsection{Stage 2: Safe Image Generation}
\label{sec:stage2_editing}

\subsubsection{Overall pipeline}
Stage~2 converts each unsafe image into a privacy-safe counterpart using a text-guided diffusion editor. Our pipeline is model-agnostic and supports instruction-driven editors such as InstructPix2Pix~\cite{brooks2023instructpix2pix} and FlowEdit~\cite{kulikov2025flowedit}. The goal is to produce images that satisfy three criteria:  
(i) sensitive attributes are neutralized or replaced,  
(ii) non-private semantics and spatial layout are preserved, and  
(iii) the result remains useful for downstream tasks.

Given the \emph{public caption} $c^{pub}$ and the \emph{edit instruction} $c^{edit}$ from Stage~1, the editor is conditioned on both signals. The edit instruction specifies what must change, while the public caption anchors the model to the safe, task-relevant description of the scene. However, standard diffusion editors attend overwhelmingly to the instruction, which can cause the model to overwrite non-sensitive regions and lose details important for downstream utility.
\input{figures/safe_attention}

\subsubsection{Safe Cross Attention}
Modern instruction-based editing models tend to prioritize making changes rather than preserving non-sensitive content. Because the source image alone serves as the only structural anchor, these models often over-apply edits or unintentionally modify regions that should remain untouched, an undesirable behavior for privacy-preserving anonymization. To address this limitation, we introduce \textbf{Safe Cross Attention}, an auxiliary attention module that conditions the denoising process jointly on the public caption $c^{\text{pub}}$ and the edit instruction $c^{\text{edit}}$.

An overview of the new module and its placement in the transformer framework is provided in Figure~\ref{fig:safeattn_unet}. Safe Cross Attention mirrors the architecture of standard cross-attention but is extended to operate on a concatenated token sequence: we combine the embeddings of $c^{\text{pub}}$ and $c^{\text{edit}}$ into a unified sequence, which is then projected into Key and Value tensors. This design provides the model with two complementary signals during denoising: (1) semantic preservation, enforced by the public caption, and (2) targeted transformation, directed by the edit instruction. By having simultaneous access to both forms of conditioning, the model can preserve layout and non-private objects while applying strong, localized edits to sensitive regions.

%% file: figures/pipeline.tex
\begin{figure}[t]
\centering
\includegraphics[trim=0pt 380pt 0pt 120pt, clip,width=0.98\linewidth]{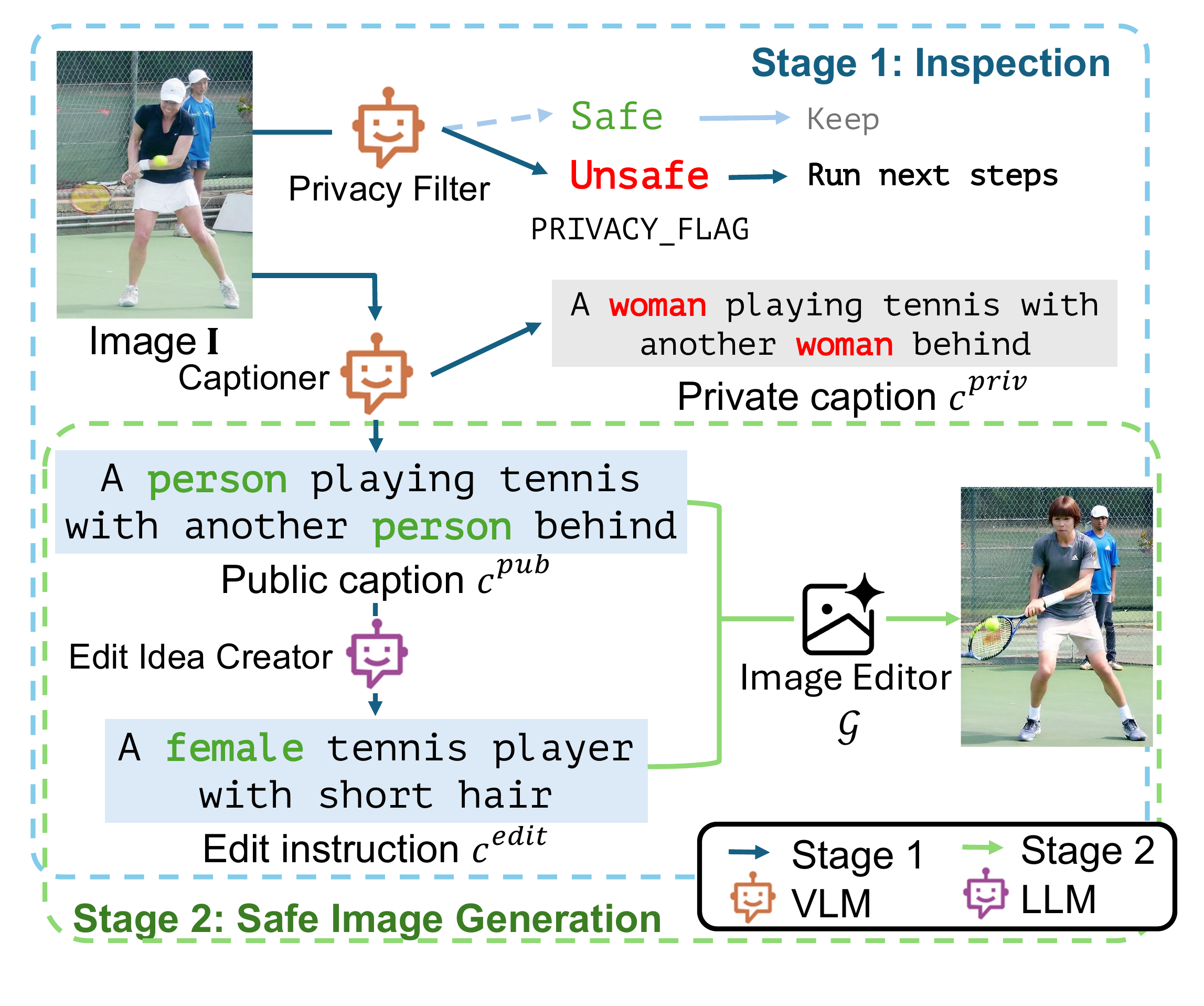}\\
\caption{\textbf{Pipeline Overview.} A VLM inspects the image for privacy risks. For flagged images, it generates a private caption $c^{priv}$ and a public caption $c^{pub}$ without sensitive details. An LLM then produces an edit instruction $c^{edit}$ on how sensitive attributes should be modified. A diffusion editor uses these priors to generate a privacy-safe image while preserving scene semantics.}
\vspace{-0.15in}
\label{fig:pipeline}
\end{figure}

%% file: figures/safe_attention.tex
\begin{figure}[t]
\centering
\vspace{-5pt}

\includegraphics[width=0.9\linewidth]{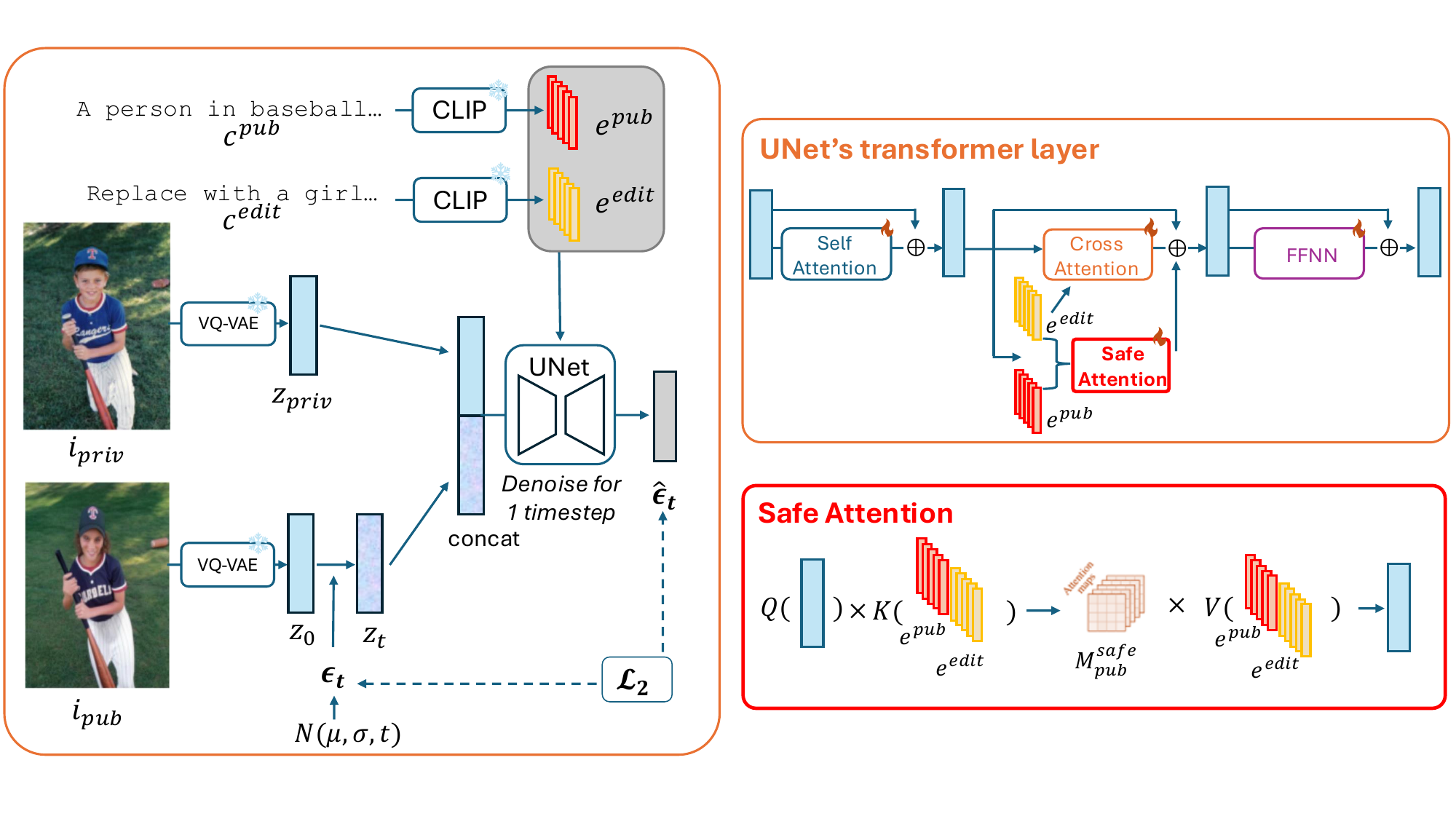}\\

\caption{\textbf{ SafeAttention within UNet.}
The UNet transformer receives two textual conditions: the edit instruction and the public caption.
The Cross Attention module follows the standard cross-attention pathway, while an auxiliary \textbf{Safe Attention} module operates on both embeddings to reinforce non-private semantics during denoising.}

\vspace{-10pt}
\label{fig:safeattn_unet}
\end{figure}

%% file: sec/4_results.tex
\section{Results}
In this section, we present a comprehensive evaluation of our \sysname{} pipeline across privacy, utility, cheating, and realism metrics. We compare against strong anonymization baselines, analyze tradeoffs between privacy and downstream performance, and examine the effects of model variants and fine-tuning. These experiments demonstrate that \sysname{} provides remarkably stronger privacy protection while maintaining competitive task utility.

\subsection{Datasets}

\noindent\textbf{Evaluation.}
For downstream evaluation, we use multiple  benchmarks: 
Caltech101~\cite{caltech101} for object classification, MIT Indoor67~\cite{MITindoor67} for scene classification, and  MS-COCO~\cite{lin2014microsoft} for image captioning and open-ended VQA based on the labels from OK-VQA~\cite{okvqa}. Though not originally designed for privacy analysis, many images contain privacy-sensitive content unrelated to their annotated categories. While prior work typically uses the first two datasets only for classification, we are the first to employ them for evaluating anonymization and privacy preservation. 

\noindent\textbf{Training Data for Anonymization.} As existing anonymization datasets are limited either in scale or in the range of identity attributes they cover, we construct our own dataset to train a generalizable privacy-preserving editor. We use MS-COCO~\cite{lin2014microsoft} (train2014 and val2014) as a large and diverse source of images and generate an edited counterpart for each image using~\cite{Liu_2024_CVPR}, which preserves scene structure while altering subject appearance. We follow the recommended configuration from the original work and set the self-attention swapping probability to 0.4 to induce pronounced visual changes. Edits that fail to preserve the original scene semantics are discarded; details of this filtering procedure are provided in Appendix~\ref{sec:coco_filtering}.

\subsection{Baselines \& Model Variants} 

\textbf{Image Anonymizer.}
We compare our approach against two recent anonymization methods, 
DeepPrivacy2~\cite{hukkelaas2023deepprivacy2} and {FaceAnon}~\cite{kung2025face}, 
both of which detect and anonymize faces and bodies. To demonstrate the adaptability of our  \sysname pipeline, we experiment with a broad set of widely used image-editing models for performing anonymization from both UNet and DiT backbone families. For UNet editors, we evaluate FreePrompt~\cite{Liu_2024_CVPR} and InstructPix2Pix~\cite{brooks2023instructpix2pix}. For DiT editors, we use FlowEdit~\cite{kulikov2025flowedit}, the current state-of-the-art in image editing, built on Stable Diffusion 3~\cite{esser2024scaling} backbone. We further fine-tune InstructPix2Pix~\cite{brooks2023instructpix2pix}. All InstructPix2Pix experiments are initialized from the pretrained MagicBrush weights~\cite{zhang2023magicbrush}.
% We further fine-tune InstructPix2Pix~\cite{brooks2023instructpix2pix} and FLUX using our anonymization dataset via OmniControl~\cite{tan2025ominicontrol}.
 
\noindent\textbf{Text Priors.}
We evaluate five types of textual priors produced in Stage~1 of our pipeline:
$c_{\text{priv}}$, \texttt{class}, $c_{\text{public}}$, $c_{\text{edit}}$, and an
LLM-composed combination of $(c_{\text{edit}}, c_{\text{public}})$.
The \texttt{class} prompt provides a minimal class-level description 
\prompt{A realistic image of <class>.}
to anchor the global concept.
The public caption $c_{\text{public}}$ contains a sanitized description of the
source image, while $c_{\text{edit}}$ specifies where and how privacy-sensitive regions should be modified.
In all experiments, we used Qwen3-4B-Instruct \cite{yang2025qwen3} as the LLM  and InternVL2.5 as the VLM agent for text prior generation. {See Appendix~\ref{sec:supp_promts_stage1} for the exact prompts.}

\subsection{Evaluation Metrics}
\label{sec:metrics}

To comprehensively assess anonymization performance, we introduce a unified evaluation framework consisting of four metric groups: \emph{Quality}, \emph{Cheating}, \emph{Privacy}, and \emph{Utility} scores. These metrics quantify realism, unintended information leakage, demographic obscurity, and downstream task performance, respectively.

\medskip
\noindent\textbf{Quality Score.}
We evaluate visual realism via CLIP similarity~\cite{radford2021learning} between each anonymized image and its public caption. Higher alignment indicates better preservation of the original scene's global semantic concept.

\medskip
\noindent\textbf{Cheating Scores.}
To measure unintended copying of pixel-level or perceptual cues, we compute (i) structural similarity (SSIM)~\cite{1284395} and (ii) perceptual similarity (LPIPS)~\cite{zhang2018perceptual}.  Lower SSIM and higher LPIPS indicate greater deviation from the source image. We refer to these as \emph{cheating scores} because anonymizers should not rely on reconstructing original identity features to maintain realism.

\medskip
\noindent\textbf{Privacy Scores.}
We propose four privacy-focused metrics derived from VLM-based analysis:

\begin{itemize}[leftmargin=1.0em]
    \item  {\textbf{VLMScore ($\uparrow$):} A VLM receives the unsafe-safe image pair and assigns a score (0--100) indicating how effectively privacy-sensitive issues have been resolved.}
    \item \textbf{FaceSim ($\downarrow$):} Cosine similarity between detected faces and their closest anonymized counterparts using the Antevelop-v2 encoder~\cite{deepinsight_2023}.
    \item \textbf{TextSim ($\downarrow$):} We extract text from each anonymized image via a VLM and compute a token-set ratio, where lower similarity indicates successful removal or distortion of identifiable textual content.
    \item \textbf{Race Entropy ($\uparrow$):} A VLM predicts demographic attributes from the set $\mathcal{R}$ present in each image (White, Black, Asian, Hispanic, and/or Other), and we compute the normalized entropy:
\[
P(r) = \frac{\mathrm{count}(r)}{\sum_{r' \in \mathcal{R}} \mathrm{count}(r')},
\]
\[
e = \frac{-\sum_{r \in \mathcal{R}} P(r)\log P(r)}{\log K}.
\]
where $K = |\mathcal{R}|$ is the number of race categories. Higher entropy indicates a more uniform, and thus less identity-specific, demographic distribution.
\end{itemize}
We refer to Appendix~\ref{sec:supp_promts_eval} for each task's VLM prompts. Privacy scores are reported on all training samples for Caltech101 and Indoor67, and only on training samples flagged as private by our VLM for MS-COCO.

\medskip
\noindent\textbf{Utility Score.} { To measure task usefulness, we fine-tune models on a wide range of downstream tasks, including object classification, scene classification, image captioning, and open-ended VQA.}

% \noindent
For classification, we train a classifier on each anonymized training set and report top-1 accuracy. For image captioning, we fine-tune a VLM~\cite{blip2} using the human-annotated MS-COCO captions and evaluate performance using BLEU-4~\cite{bleuscore} and CIDEr~\cite{Vedantam_2015_CVPR}, measuring alignment between generated captions and human references. For VQA, we also fine-tune a VLM~\cite{yang2025qwen3} and report VQA accuracy~\cite{VQA_paper}. 
Importantly, all utility scores are computed on the \emph{original} test sets, reflecting the degree of task-relevant semantics preservation under the original data distribution rather than adaptation to anonymized data.

\noindent

\medskip

These four metric groups provide a holistic evaluation of anonymization quality, information leakage, demographic privacy, and practical utility, representing a key contribution of our work to assess anonymization methods. More details on the implementation of these evaluation metrics are provided in Appendix~\ref{sec:eval_details}.

\subsection{Main Results}
\input{tables/vispr_results}

\paragraph{Is Stage~1 reliable?}
We assess the reliability of the Stage~1 privacy inspection module on VISPR~\cite{orekondy17iccv}, treating the task as binary privacy detection. 
Since missed detections directly translate to privacy leakage in the training set, recall is the most critical metric. As shown in Table~\ref{tab:privacy_metrics_core}, the detector achieves consistently high recall across privacy categories, indicating that only a negligible fraction of sensitive samples remain unflagged. Implementation details and attribute group definitions are provided in Appendix~\ref{appex:stage1_details}. 

\input{figures/realism_exp}
\input{tables/quality_table}
\paragraph{Does the Model Generate Realistic and Properly Anonymized Images?}
We examine whether \sysname{} produces visually coherent edits while effectively removing privacy-sensitive attributes. Figure~\ref{fig:realism_examples} shows qualitative results comparing prior privacy-preserving methods and our pipeline. All models preserve the global layout and structure of the original scene while replacing the identities of the person in the image. Importantly, \sysname also anonymizes background elements and non-human objects (i.e. posters and ads on the board) whenever they contain privacy-unsafe cues—an ability largely absent in prior approaches, which tend to modify only facial regions. 

This observation is corroborated quantitatively in Table~\ref{tab:quality_cheating}. CLIP scores remain comparable across methods, indicating similar perceptual realism. Yet, \sysname{} consistently yields lower SSIM and higher LPIPS values, revealing stronger privacy-preserving transformations without compromising global scene semantics. The margins of change are smaller on Caltech101 due to its higher proportion of publicly available, non-sensitive images.

\input{tables/utility_privacy_table}
\input{tables/mscoco_results}
\input{tables/mscoco_okvqa_results}

\vspace{-0.5cm}
\paragraph{Trade-Off Between Utility and Privacy.}
To evaluate how much privacy protection can be achieved without sacrificing downstream performance, we analyze the trade-off between utility—measured by classification accuracy, captioning scores (BLEU-4 and CIDEr), and VQA accuracy—and privacy, assessed using VLMScore, FaceSim, TextSim, and Race Entropy.

\vspace{0.1cm}
\noindent\textbf{Utility.}  
As shown in Table~\ref{tab:utility_privacy_tradeoff}, classifiers trained on U2S-anonymized datasets achieve performance close to models trained on the original images, with differences as small as 0.5 points on Caltech101 and 4.6 points on Indoor67. A larger drop occurs only when anonymization relies on FlowEdit with the \textit{edit} instruction, likely due to weaker alignment between the instruction formulation and dataset captions.

A similar pattern appears in generative tasks. On MS-COCO captioning (Table~\ref{tab:mscoco_results}), BLEU-4 and CIDEr remain near the raw baseline despite strong privacy transformations. On OK-VQA (Table~\ref{tab:okvqa_results}), models trained on U2S-generated images achieve accuracy comparable to, and in our experiments slightly exceeding, those trained on raw or face-anonymized data. These results indicate that the anonymization process preserves task-relevant semantics required for language grounding and reasoning.

\input{tables/race_distribution}
\input{tables/comparison_with_finetuned}

\vspace{0.1cm}
\noindent\textbf{Privacy.} 
\sysname(U2S) consistently delivers stronger privacy protection across all metrics. It achieves substantially higher \textit{VLMScore}, and thus stronger removal of instance-level identity cues, as evaluated by the VLM judge. This suggests that anonymization is not limited to pixel-level modifications but effectively disrupts semantic identity linkage. In all datasets, U2S significantly reduces \textit{face similarity} compared to existing anonymizers, achieving up to 5 points on Caltech101, 15 points on Indoor67, and 24 points on the captioning and VQA training sets. \textit{TextSim} values are likewise markedly lower, showing that identifiable textual content is frequently altered or rendered unreadable.  Although Caltech101 exhibits a highly skewed demographic distribution, U2S introduces noticeably \textit{race entropy} (i.e., higher diversity). Table~\ref{tab:race_distribution} provides the breakdown of race distributions for all models. Additional experiments demonstrate that demographic attributes can be further controlled through textual conditioning (Appendix~\ref{appdx:controlability}).

\noindent
Overall, \sysname preserves downstream performance while providing stronger privacy protection.

\begin{figure}[h]
\centering
\includegraphics[width=0.9\linewidth,trim=50pt 50pt 50pt 20pt,clip]{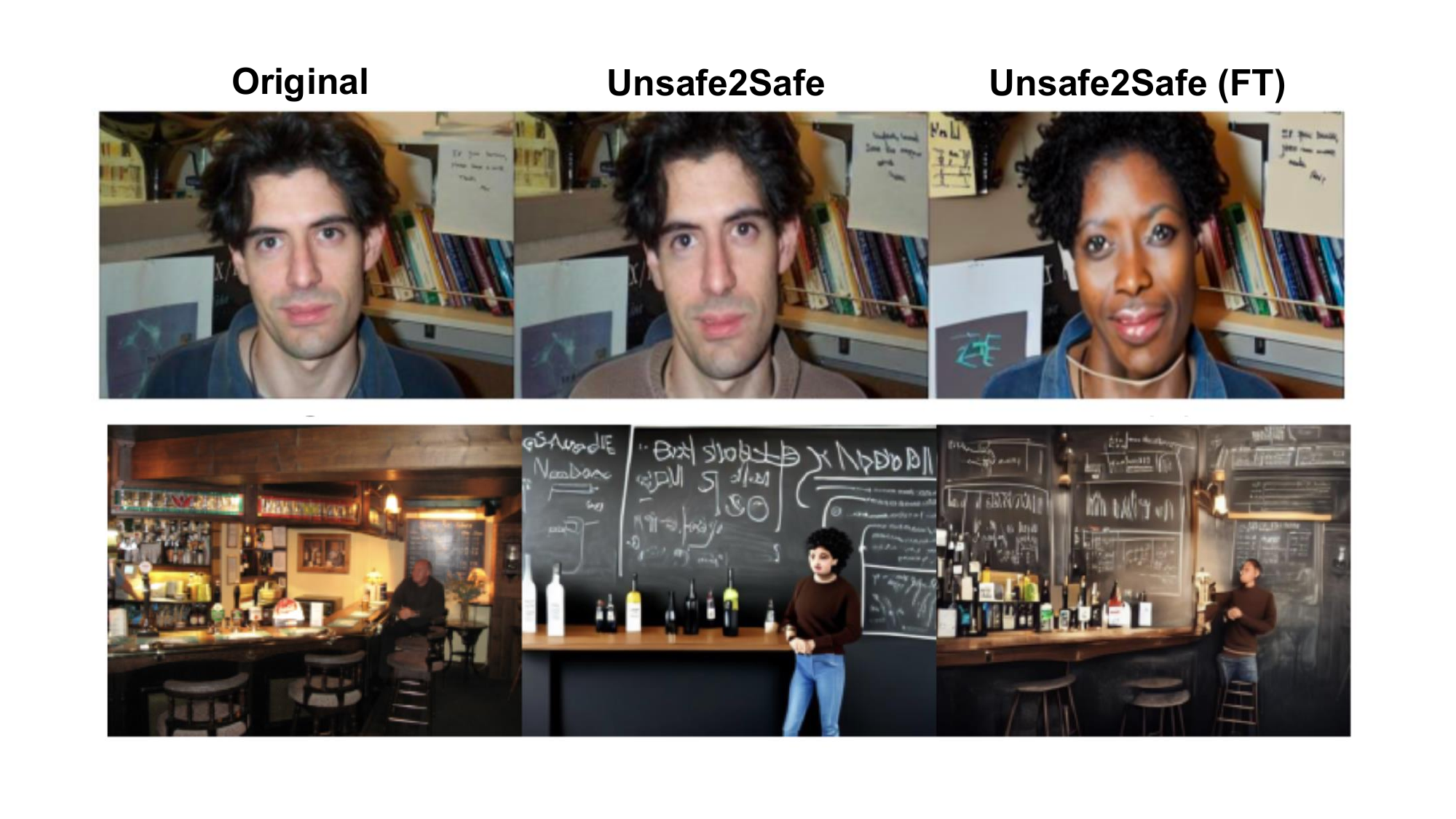}
\caption{
Qualitative examples showing that the  InstructPix2Pix model finetuned on our dataset
(\textbf{FT}) more effectively anonymizes sensitive content while preserving original class semantics, compared to the model trained on general editing data.
}
\vspace{-0.5cm}
\label{fig:ft_success_cases}
\end{figure}

\paragraph{Is Fine-Tuning and SafeAttention Helpful?}
We further evaluate whether fine-tuning InstructPix2Pix on our Unsafe2Safe dataset improves utility and privacy performance. Table~\ref{tab:instructpix2pix_ft} reports quantitative results before and after fine-tuning, and Figure~\ref{fig:ft_success_cases} shows representative examples. Under the same $c_{\text{edit}}$ prior, the fine-tuned model achieves more effective anonymization than its non-fine-tuned counterpart, better removing facial identity cues while preserving scene composition (e.g., the bar layout). In the top example, the non-fine-tuned Unsafe2Safe output leaves residual identity cues in the face. This observation is consistent with the general-purpose design of InstructPix2Pix, and highlights the benefit of task-specific fine-tuning on our dataset.

Fine-tuning provides clear benefits. The fine-tuned $c_{\text{edit}}$ model achieves the highest Caltech101 accuracy (95.116\%) among all InstructPix2Pix variants, while also 
improving privacy metrics such as FaceSim and TextSim on Indoor67. Notably, the fine-tuned models also exhibit stronger race entropy, indicating a more diverse and less identity-specific demographic distribution compared to their non-finetuned counterparts. These results show that fine-tuning helps adapt the editor to the anonymization task, improving both utility and privacy outcomes.

The SafeAttention variant further improves anonymization stability. Compared to the non-finetuned and standard finetuned models, the SafeAttention model achieves the strongest overall privacy protection, including the lowest Indoor FaceSim and the highest Caltech and Indoor race entropy as well as Indoor VLMScore. These gains come with no noticeable loss in utility, demonstrating that SafeAttention helps guide the editor toward safer and more demographic-neutral outputs. See Appendix~\ref{appdx:safe_attn_maps} for more analysis on the attention maps about guidance from $c_{\text{public}}$.

% \medskip
% \noindent\textbf{Anonymization Patterns.} \\

%% file: tables/vispr_results.tex
\begin{table}[t]
\centering
\caption{VLM privacy detector performance under different attribute flagging criteria; higher recall means less privacy leakage. }

\resizebox{0.8\columnwidth}{!}{\begin{tabular}{l|cccc}

\toprule

\textbf{Privacy Criterion} & \textbf{Recall} $\uparrow$ & \textbf{Precision} $\uparrow$ & \textbf{F1} $\uparrow$ \\
\midrule
All attributes & 0.975 & 0.793 & 0.874 \\
Face & 0.850 & 0.927 & 0.887 \\
Health indicators & 0.892 & 0.678 & 0.770 \\
Vehicles & 0.829 & 0.435 & 0.570 \\
Personal opinion & 0.778 & 0.665 & 0.717 \\
\bottomrule
\end{tabular}}

\label{tab:privacy_metrics_core}
\end{table}

%% file: figures/realism_exp.tex
\begin{figure}[t]
\centering
\vspace{-5pt}
\includegraphics[width=0.9\linewidth,trim=10pt 50pt 50pt 50pt,clip]{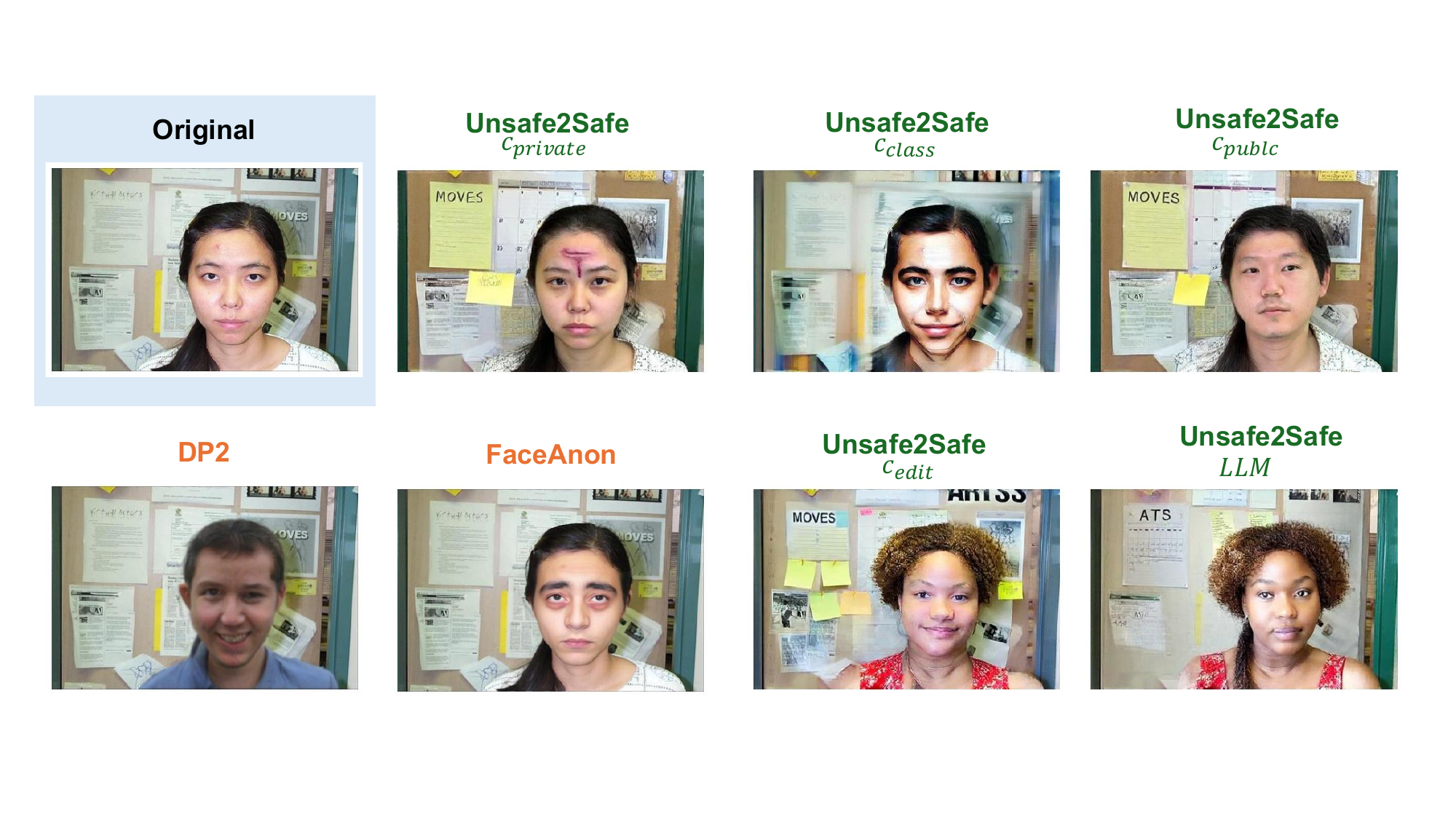}
\caption{\textbf{Qualitative comparison of anonymization outputs on Caltech101.}
Each image shows with a different model family (\textbf{top line}) and its textual condition (\textit{bottom line}). 
All methods preserve the global layout of the original scene, but the Unsafe2Safe using FlowEdit~\cite{kulikov2025flowedit}, unlike face-only anonymizers (DP2, FaceAnon), modify background elements when they contain privacy-relevant cues, while keeping overall scene composition intact.}
\label{fig:realism_examples}

\end{figure}

%% file: tables/quality_table.tex
\begin{table}[t]
\centering
\small
\setlength{\tabcolsep}{3.5pt}
\renewcommand{\arraystretch}{1.15}

\caption{
\textbf{Quality and Cheating scores for Caltech101 (Cal101) and MIT Indoor 67 (I67).} Our \sysname{} (U2S), which uses FlowEdit (FE)~\cite{kulikov2025flowedit} as the underlying image editor, generates high-quality images comparable to other anonymization models. Unlike prior methods that rarely modify content beyond facial regions, U2S identifies privacy-unsafe cues throughout the entire image and performs broader, context-aware edits. Best values are highlighted in yellow, and second-best in light blue.
}

\resizebox{\columnwidth}{!}{
\begin{tabular}{ll|cc|cc|cc}
\toprule
\multirow{3}{*}{\textbf{Model}} &
\multirow{3}{*}{\textbf{Prior}} &
\multicolumn{2}{c|}{\textbf{Quality Score}} &
\multicolumn{4}{c}{\textbf{Cheating Scores}} \\
\cmidrule(lr){3-4}\cmidrule(lr){5-6}\cmidrule(lr){7-8}
& & \multicolumn{2}{c|}{CLIP (↑)} & \multicolumn{2}{c|}{SSIM (↓)} & \multicolumn{2}{c}{LPIPS (↑)} \\
\cmidrule(lr){3-4}\cmidrule(lr){5-6}\cmidrule(lr){7-8}
& & C101 & I67 & C101 & I67 & C101 & I67 \\
\midrule

\multirow{5}{*}{\textbf{U2S (FE)}}
& $c_{\text{priv}}$ 
   & 0.3054 
   & 0.3411 
   & \cellcolor{secondcell}\textbf{0.8551} 
   & \cellcolor{secondcell}\textbf{0.6497} 
   & 0.1118 
   & 0.2815 \\

 & class              
   & 0.2634 
   & 0.3149 
   & 0.8559 
   & 0.6708 
   & \cellcolor{secondcell}\textbf{0.1139} 
   & 0.2600 \\

 & $c_{\text{pub}}$   
   & \cellcolor{bestcell}\textbf{0.3094} 
   & 0.3448 
   & 0.8556 
   & \cellcolor{secondcell}\textbf{0.6497} 
   & 0.1118 
   & 0.2803 \\

 & $c_{\text{edit}}$  
   & 0.2685 
   & 0.2946 
   & \cellcolor{bestcell}\textbf{0.8550} 
   & 0.6580 
   & \cellcolor{bestcell}\textbf{0.1168} 
   & \cellcolor{secondcell}\textbf{0.2846} \\

 & LLM                
   & 0.2865 
   & 0.3251 
   & \cellcolor{secondcell}\textbf{0.8551} 
   & \cellcolor{bestcell}\textbf{0.6484} 
   & \cellcolor{bestcell}\textbf{0.1168} 
   & \cellcolor{bestcell}\textbf{0.2896} \\
\midrule

DP2 
 & -- 
   & 0.3023 
   & \cellcolor{bestcell}\textbf{0.3512} 
   & 0.9817 
   & 0.9608 
   & 0.0161 
   & 0.0365 \\

FaceAnon
 & -- 
   & \cellcolor{secondcell}\textbf{0.3057} 
   & \cellcolor{secondcell}\textbf{0.3452} 
   & 0.9888 
   & 0.9443 
   & 0.0104 
   & 0.0812 \\
\bottomrule
\end{tabular}
}
\label{tab:quality_cheating}
\end{table}

%% file: tables/utility_privacy_table.tex
\begin{table*}[th]
\centering
\footnotesize
\setlength{\tabcolsep}{6pt}
\renewcommand{\arraystretch}{1.15}
\caption{
\textbf{Utility and privacy comparison on classification tasks.} 
Our \sysname, which uses FlowEdit (FE) \cite{kulikov2025flowedit} as the underlying image editor, 
achieves comparable performance on downstream tasks while successfully anonymizing 
privacy-sensitive information, unlike other anonymization models. 
The \textbf{best} value (yellow) and the \textbf{second-best} value (blue) are highlighted per column. 
}
\begin{tabular}{ll|cc|cc|cc|cc|cc}
\toprule
\multirow{3}{*}{\textbf{Model}} &
\multirow{3}{*}{\textbf{Text Prior}} &
\multicolumn{2}{c|}{\textbf{Utility Score}} &
\multicolumn{8}{c}{\textbf{Privacy Score}} \\
\cmidrule(lr){3-4}
\cmidrule(lr){5-12}
& &
\multicolumn{2}{c|}{\textbf{Accuracy ($\uparrow$)}} &
\multicolumn{2}{c|}{\textbf{VLMScore ($\uparrow$)}} &
\multicolumn{2}{c|}{\textbf{FaceSim ($\downarrow$)}} &
\multicolumn{2}{c|}{\textbf{TextSim ($\downarrow$)}} &
\multicolumn{2}{c}{\textbf{Race Entropy ($\uparrow$)}} \\
& &
Cal101 & Indoor &
Cal101 & Indoor &
Cal101 & Indoor &
Cal101 & Indoor &
Cal101 & Indoor \\
\midrule

% ---------------- Raw images ----------------
Raw Images & -- &
94.277 & 83.881 &
7.700&0.763&
1.0000 & 1.0000 &
1.0000 & 1.0000 &
0.4384 & 0.7443 \\
\midrule

% ---------------- FlowEdit block ----------------
\multirow{5}{*}{\textbf{\sysname (FE)}}
 & $c_{\text{private}}$ &
 
94.334 & 79.925 &
9.646&8.009&
0.4378 & 0.2666 &
0.6611 & 0.4210 &
{0.5552} & 0.7399 \\

 & class &
93.857 & 80.448 &
12.555&7.512&
0.4965 & 0.3743 &
\cellcolor{bestcell}{\textbf{0.4856}} & \cellcolor{bestcell}{\textbf{0.2077}} &
{0.4051} & 0.6508 \\

 & $c_{\text{public}}$ &
94.487 & 80.746 &
9.873&7.440&
0.4881 & 0.2883 &
0.5395 & 0.2896 &
{0.6409} & 0.7208 \\

 & $c_{\text{edit}}$ &
\cellcolor{secondcell}{\textbf{94.792}} & 77.090 &
\cellcolor{bestcell}\textbf{13.966}&\cellcolor{bestcell}\textbf{21.390}
&
\cellcolor{secondcell}{\textbf{0.3658}} & \cellcolor{bestcell}{\textbf{0.2077}} &
0.5238 & 0.2393 &
\cellcolor{secondcell}{\textbf{0.7646}} & \cellcolor{secondcell}{\textbf{0.7589}} \\

 & LLM ($c_{\text{edit}},c_{\text{public}}$) &
92.884 & 80.746 &
\cellcolor{secondcell}\textbf{12.695}&\cellcolor{secondcell}\textbf{14.937}&
\cellcolor{bestcell}{\textbf{0.3428}} & \cellcolor{secondcell}{\textbf{0.2294}} &
\cellcolor{secondcell}{\textbf{0.4881}} & \cellcolor{secondcell}{\textbf{0.2119}} &
\cellcolor{bestcell}{\textbf{0.8751}} & \cellcolor{bestcell}{\textbf{0.7643}} \\
\midrule

% ---------------- Baselines ----------------
DeepPrivacy2~\cite{hukkelaas2023deepprivacy2} & -- &
94.601 & \cellcolor{bestcell}{\textbf{84.030}} &
11.053&0.767&
0.3921 & 0.3547 &
0.9569 & 0.8653 &
0.7315 & 0.7547 \\

FaceAnonSimple~\cite{kung2025face} & -- &
\cellcolor{bestcell}{\textbf{94.849}} & \cellcolor{bestcell}{\textbf{84.030}} &
8.757&1.233&
0.4586 & 0.5045 &
0.9355 & 0.7701 &
0.6091 & 0.7407 \\

\bottomrule
\end{tabular}

\label{tab:utility_privacy_tradeoff}
\end{table*}

%% file: tables/mscoco_results.tex
\begin{table}[t]
\centering
\footnotesize
\setlength{\tabcolsep}{1.2pt}

\caption{
\textbf{Utility and privacy comparison on MS-COCO captioning.}
U2S achieves higher VLMScore and much lower FaceSim/TextSim. Our \sysname uses FreePrompt (FP)~\cite{Liu_2024_CVPR} and FlowEdit (FE) \cite{kulikov2025flowedit} as the underlying image editor.
% \textbf{Best} (yellow) and \textbf{second-best} (blue) values are highlighted per column.
}
\label{tab:mscoco_results}

\resizebox{\linewidth}{!}{%
\begin{tabular}{ll|cc|cccc}
\toprule

\multirow{2}{*}{\textbf{Model}} &
\multirow{2}{*}{\makecell{\textbf{Text}\\\textbf{Prior}}} &
\multicolumn{2}{c|}{\textbf{Utility}} &
\multicolumn{4}{c}{\textbf{Privacy}} \\
\cmidrule(lr){3-4}
\cmidrule(lr){5-8}

&&
BLEU-4 $\uparrow$ & CIDEr $\uparrow$ &
VLMScore $\uparrow$ & FaceSim $\downarrow$ &
TextSim $\downarrow$ & RaceEnt $\uparrow$ \\
\midrule

Raw & -- &
\cellcolor{secondcell}\textbf{0.436} &
\cellcolor{secondcell}\textbf{1.413} &
0.646 & 1.0000 & 1.0000 & 0.6170 \\
\midrule

U2S (FP) & $c_{\text{edit}}$ &
0.433 & 1.390 &
\cellcolor{secondcell}\textbf{32.002} &
\cellcolor{secondcell}\textbf{0.2013} &
\cellcolor{secondcell}\textbf{0.1442} &
\cellcolor{bestcell}\textbf{0.7364} \\

U2S (FE) & $LLM$ &
0.423 & 1.363 &
\cellcolor{bestcell}\textbf{36.641} &
\cellcolor{bestcell}\textbf{0.1975} &
\cellcolor{bestcell}\textbf{0.1276} &
\cellcolor{secondcell}\textbf{0.7282} \\
\midrule

FaceAnon & -- &
\cellcolor{bestcell}\textbf{0.444} &
\cellcolor{bestcell}\textbf{1.429} &
1.622 & 0.4436 & 0.8123 & 0.6458 \\
\bottomrule
\end{tabular}
}
\end{table}

%% file: tables/mscoco_okvqa_results.tex
\begin{table}[thb]
\centering
\footnotesize
\setlength{\tabcolsep}{1.2pt}

\caption{
\textbf{Utility and privacy comparison on OK-VQA.}
U2S attains the highest VQA accuracy while achieving the strongest privacy scores.
Best (yellow) values are highlighted per column.
}

\resizebox{\linewidth}{!}{%
\begin{tabular}{lc|c|cccc}
\toprule
\multirow{2}{*}{\textbf{Model}} &
\multirow{2}{*}{\makecell{\textbf{Text}\\\textbf{Prior}}} &
\multicolumn{1}{c|}{\textbf{Utility}} &
\multicolumn{4}{c}{\textbf{Privacy}} \\
\cmidrule(lr){3-3}
\cmidrule(lr){4-7}

&&
$\text{Acc}_{\text{VQA}}$ ($\uparrow$) &
{VLMScore ($\uparrow$)} & FaceSim ($\downarrow$) & TextSim ($\downarrow$) & RaceEnt ($\uparrow$) \\
\midrule

Raw  & -- &
0.6064 &
0.600 & 1.0000 & 1.0000 & 0.6288 \\
\midrule

U2S (FP) & $c_{\text{edit}}$ &
0.6573 &
\cellcolor{secondcell}\textbf{33.192} & \cellcolor{secondcell}\textbf{0.2041 }& \cellcolor{secondcell}\textbf{0.1514} & \cellcolor{bestcell}\textbf{0.7499} \\

U2S (FE) & $LLM$ &
\cellcolor{bestcell}\textbf{0.7093} &
\cellcolor{bestcell}\textbf{37.059} & \cellcolor{bestcell}\textbf{0.1951} & \cellcolor{bestcell}\textbf{0.1196} & \cellcolor{secondcell}\textbf{0.7397} \\
\midrule

FaceAnon  & -- &
\cellcolor{secondcell}\textbf{0.6632} &
1.705 & 0.4483 & 0.8158 & 0.6568 \\
\bottomrule
\end{tabular}
}
\label{tab:okvqa_results}
\end{table}

%% file: tables/race_distribution.tex
\begin{table}[h]
\centering
\caption{Race distribution (\%) of images of different models.}
\setlength{\tabcolsep}{2pt}   % default is 6pt
\renewcommand{\arraystretch}{0.95} % reduce row height slightly
\resizebox{\columnwidth}{!}{%
\begin{tabular}{l|ccccc}
\toprule
\textbf{Model} & \textbf{\%White} & \textbf{\%Black} & \textbf{\%Asian} & \textbf{\%Hispanic} & \textbf{\%Other} \\
\midrule
Raw Images & 80.28 & 2.82 & 5.63 & 4.23 & 7.04 \\
\midrule
U2S (FE, $c_{\text{edit}}$) & 45.93 & 29.63 & 13.33 & 0.00 & 11.11 \\
U2S (FE, $LLM$) & 37.90 & 25.81 & 17.74 & 2.42 & 16.13 \\
\midrule
DP2 \cite{hukkelaas2023deepprivacy2} & 56.70 & 3.09 & 19.59 & 4.12 & 16.49 \\
FaceAnon \cite{kung2025face} & 70.67 & 4.00 & 10.67 & 4.00 & 10.67 \\
\bottomrule
\end{tabular}}
\label{tab:race_distribution}
\end{table}

%% file: tables/comparison_with_finetuned.tex
\begin{table*}[t!]
\centering
\footnotesize
\setlength{\tabcolsep}{6pt}
\renewcommand{\arraystretch}{1.15}

\caption{
Performance of InstructPix2Pix~\cite{brooks2023instructpix2pix} with different text priors, 
and after fine-tuning on our anonymization dataset (\ding{51}). 
Fine-tuning notably improves both utility and privacy: the fine-tuned 
$c_{\text{edit}}$ model achieves the highest Caltech101 accuracy (95.116\%) among all variants, 
while also reducing FaceSim/TextSim on Indoor67 and increasing race entropy, 
indicating more diverse and less identity-specific outputs. 
\textcolor{yesgreen}{\textbf{Safe}} denotes the model trained with Safe Attention.
}

\begin{tabular}{l l |cc|cc|cc|cc|cc}
\toprule
\multirow{3}{*}{\textbf{Finetuned}} &
\multirow{3}{*}{\textbf{Text Prior}} &
\multicolumn{2}{c|}{\textbf{Utility Score}} &
\multicolumn{8}{c}{\textbf{Privacy Score}} \\
\cmidrule(lr){3-4}
\cmidrule(lr){5-12}

& &
\multicolumn{2}{c|}{\textbf{Accuracy ($\uparrow$)}} &
\multicolumn{2}{c|}{\textbf{VLMScore ($\uparrow$)}} &

\multicolumn{2}{c|}{\textbf{FaceSim ($\downarrow$)}} &
\multicolumn{2}{c|}{\textbf{TextSim ($\downarrow$)}} &
\multicolumn{2}{c}{\textbf{Race Entropy ($\uparrow$)}} \\

& &
Cal101 & Indoor &
Cal101 & Indoor &
Cal101 & Indoor &
Cal101 & Indoor &
Cal101 & Indoor \\
\midrule

% ===========================================================
% InstructPix2Pix (non-FT)
% ===========================================================

\multirow{5}{*}{}
 & $c_{\text{private}}$ &
94.773 & 80.746 &
11.651&6.559&
0.5571 & 0.3382 &
0.6553 & 0.3112 &
0.5642 & 0.7414 \\

 & class &
\cellcolor{secondcell}{\textbf{94.887}} & 79.776 &
10.674&5.943&
0.6888 & 0.3500 &
0.6176 & 0.2898 &
0.5528 & 0.7678 \\

 & $c_{\text{public}}$ &
94.582 & \cellcolor{bestcell}{\textbf{81.791}} &
10.814&5.646&
0.5401 & 0.3389 &
0.6589 & 0.3044 &
0.6497 & 0.7599 \\

 & $c_{\text{edit}}$ &
94.315 & 81.418 &
\cellcolor{bestcell}\textbf{16.878}&\cellcolor{secondcell}\textbf{16.926}&
0.5164 & 0.2909 &
\cellcolor{secondcell}{\textbf{0.6126}} & 0.3399 &
0.6826 & 0.7493 \\

 & LLM ($c_{\text{edit}},c_{\text{public}}$) &
93.762 & 81.119 &
\cellcolor{secondcell}\textbf{13.699}&12.754&
\cellcolor{bestcell}{\textbf{0.4792}} & 0.2857 &
0.6455 & 0.3112 &
0.7746 & 0.8090 \\
\midrule

% ===========================================================
% InstructPix2Pix (FT)
% ===========================================================

\ding{51} &
$c_{\text{edit}}$ &
\cellcolor{bestcell}{\textbf{95.116}} & 
\cellcolor{secondcell}{\textbf{81.716}} &
12.595&16.250&
0.5910 & 
\cellcolor{secondcell}{\textbf{0.2735}} &
0.6176 & 
\cellcolor{bestcell}{\textbf{0.2640}} &
0.7997 & 
\cellcolor{secondcell}{\textbf{0.8356}} \\

\ding{51} &
LLM ($c_{\text{edit}},c_{\text{public}}$) &
94.601 & 
81.493 &
12.145&10.804&
\cellcolor{secondcell}{\textbf{0.5150}} & 
0.2970 &
0.6464 & 
0.3090 &
\cellcolor{bestcell}{\textbf{0.8308}} & 
0.8090\\

\midrule
\ding{51} \textcolor{yesgreen}{\textbf{Safe}}&
$c_{\text{edit}},c_{\text{public}}$	&
\cellcolor{secondcell}{\textbf{94.887}}	& 80.149 &
13.366&\cellcolor{bestcell}\textbf{17.801}& 
0.5468 &
\cellcolor{bestcell}{\textbf{0.2465}}	& 
\cellcolor{bestcell}{\textbf{0.5966}} & \cellcolor{secondcell}{\textbf{0.2660}}	& 
\cellcolor{secondcell}{\textbf{0.8306}} & 	\cellcolor{bestcell}{\textbf{0.8448}} \\
\bottomrule
\end{tabular}
\label{tab:instructpix2pix_ft}
\end{table*}

%% file: sec/5_conclusion.tex
\section{Conclusion}

We presented Unsafe2Safe, a fully automated framework for transforming privacy-prone images into privacy-safe yet semantically faithful counterparts. By combining VLM-based inspection, public/private captioning, and LLM-generated edit instructions with diffusion-based editors, the system rewrites sensitive regions while preserving global structure and task-relevant semantics. Privacy criteria can be specified via textual prompts, and components can be instantiated with different VLMs and LLMs.

We introduced a unified evaluation suite spanning image quality, leakage, privacy attributes, and downstream utility, enabling holistic assessment of anonymization methods. Experiments show that Unsafe2Safe significantly reduces identity leakage while maintaining downstream accuracy, supporting scalable privacy-aware dataset construction.

Finally, Unsafe2Safe is a configurable dataset construction tool, not an autonomous arbiter of privacy. Responsibility for defining privacy criteria and deployment constraints rests with practitioners.

%% file: sec/6_acknowledgments.tex
\section*{Acknowledgments}
This work was supported by startup funds provided by Dartmouth College.

%% file: sec/X_suppl.tex
% % \clearpage
% % \setcounter{page}{1}

% \clearpage
% % \appendix
% % \section*{Appendix}
% \addcontentsline{toc}{section}{Appendix} % optional
%  \setcounter{section}{0} % optional
% %\renewcommand{\thesection}{A\arabic{section}} % optional

% % \clearpage
% \setcounter{page}{1}
% \addcontentsline{toc}{section}{Appendix} % optional

% \renewcommand{\thefigure}{S\arabic{figure}}
% \renewcommand{\thetable}{S\arabic{table}}
% \renewcommand{\thesection}{S\arabic{section}}
% \renewcommand{\thesubsection}{S\arabic{section}.\arabic{subsection}}
% \renewcommand{\theequation}{S\arabic{equation}}

% \setcounter{figure}{0}
% \setcounter{table}{0}
% \setcounter{equation}{0}

% \setcounter{section}{0}
% \setcounter{subsection}{0}

% % \maketitlesupplementary
% % \appendix
% % \clearpage
% \appendix

\clearpage
% \maketitlesupplementary
% \makebox[\textwidth][c]{
% \begin{tabular}{c}
% \large{
% Minh Dinh \quad SouYoung Jin}\\
% Dartmouth College\\
% Hanover, New Hampshire, USA\\
% {\tt\small \{Minh.T.Dinh.GR, SouYoung.Jin\}@dartmouth.edu}
% \end{tabular}}
\twocolumn[{%
\centering
{\Large\bfseries Unsafe2Safe: Controllable Image Anonymization for Downstream Utility\par}
\vspace{0.5em}
{\Large Supplementary Material\par}
\vspace{1.0em}

{\large Minh Dinh \quad SouYoung Jin\par}
\vspace{0.25em}
\large Dartmouth College\par
% Hanover, New Hampshire, USA\par
{\ttfamily\small \{Minh.T.Dinh.GR, SouYoung.Jin\}@dartmouth.edu\par}
\vspace{1.0em}
}]

\setcounter{page}{1}

% Section numbering
\renewcommand{\thesection}{S\arabic{section}}
\renewcommand{\thesubsection}{S\arabic{section}.\arabic{subsection}}
\renewcommand{\thesubsubsection}{S\arabic{section}.\arabic{subsection}.\arabic{subsubsection}}

% Figure / table numbering
\renewcommand{\thefigure}{S\arabic{figure}}
\renewcommand{\thetable}{S\arabic{table}}
\renewcommand{\theequation}{S\arabic{equation}}

% Reset counters
\setcounter{section}{0}
\setcounter{subsection}{0}
\setcounter{subsubsection}{0}
\setcounter{figure}{0}
\setcounter{table}{0}
\setcounter{equation}{0}

\addcontentsline{toc}{section}{Appendix}

\section{Dataset Construction and Preprocessing}

\noindent\textbf{Data usage.} All images are sourced from publicly available datasets (e.g., Caltech101~\cite{caltech101}, MIT Indoor Scenes~\cite{MITindoor67}, and MS-COCO~\cite{lin2014microsoft}) and used in accordance with their respective research usage policies.

\subsection{Dataset Choice} 

We select MS-COCO, Caltech101, and MIT Indoor67 because they provide standardized downstream utility labels while containing diverse real-world identity and contextual cues that require whole-image anonymization beyond facial regions.

Many commonly used computer vision datasets, such as CIFAR or iNaturalist, contain limited privacy-sensitive content and therefore do not meaningfully test anonymization methods. In contrast, datasets used by prior anonymization works~\cite{kung2025face,hukkelaas2023deepprivacy2} (e.g., CelebA-HQ, FFHQ) primarily evaluate re-identification rates and focus heavily on faces. While these benchmarks are suitable for measuring facial identity removal, they lack downstream task labels, making it difficult to assess the utility–privacy trade-off.

Web-scale datasets such as LAION or Flickr are highly representative of privacy-prone internet imagery and often contain caption annotations. However, their scale makes large-scale controlled anonymization experiments computationally prohibitive, particularly when multi-stage editing and evaluation are required.

We refrain from using ImageNet because it is likely included in the pretraining data of the VLMs, diffusion models, and utility backbones used in our pipeline. Such overlap could confound evaluation by introducing unintended memorization or distribution leakage. Our chosen datasets allow controlled evaluation of anonymization quality while minimizing potential pretraining bias.

\subsection{Filtering of MS-COCO for Content Preservation} 
\label{sec:coco_filtering}
Because the editing step may sometimes produce results that deviate from the original scene semantics, we apply a CLIP-based filtering step to retain only high-quality edited images.  
For each edited image $x'$, we compute its CLIP similarity to the corresponding public caption and normalize it by the CLIP similarity between the original private image $x$ and the same caption:
\begin{equation}
    s_{\text{norm}} = \frac{\text{CLIP}(c^{\text{public}}, x')} {\text{CLIP}(c^{\text{public}}, x)}.
\end{equation}
\noindent
This normalized score measures semantic preservation relative to the original image-caption alignment. 
A value close to 1 indicates that the edited image remains as semantically aligned with the public caption as the original image, whereas lower values suggest semantic degradation or unintended content changes.

\input{figures/clip_score_distr}

\noindent
Figure~\ref{fig:clip_score_distr} shows that the majority of edited samples cluster tightly around 0.9, demonstrating that our anonymization procedure largely preserves global semantics. The small left-tail corresponds to failure cases where the edit significantly alters scene composition or weakens alignment with the public caption. Such samples can introduce noisy supervision signals during training, potentially harming both downstream utility and generative stability.

\noindent
We retain only edited images with $s_{\text{norm}} > 0.7$. As shown in the distribution, this threshold falls in the lower tail and removes only a small fraction of samples—approximately 7.35\% of the original \textit{train2014} split (reducing it from 51,401 to 47,623). As a result, semantically degraded edits are excluded while preserving the vast majority of anonymized data. This filtering step improves the consistency of the training signal without meaningfully reducing dataset diversity, thereby balancing semantic fidelity and robustness in the anonymized training set. Notably, the resulting 47,623 anonymized image pairs constitute one of the largest publicly constructed before–and–after datasets for image editing~\cite{zhang2023magicbrush}, and, to our knowledge, the largest that is explicitly privacy-aware.

\section{Implementation Details}
\label{sec:supp_implementation}
\subsection{Language models}
Through out all experiments, we used InternVL2.5~\cite{chen2024expanding} as the VLM and Qwen3-4B with non-thinking mode (Qwen3-4B-Instruct-2507) \cite{yang2025qwen3} as the LLM. This choice is mainly empiracal towards a fast but robust model. Once obtaining the answers from the model, we automatically parse the sections following the structure in the prompt. 
% In the captioning steps, there were some cases when the VLM returned repeated tokens until maxing out the token limits, for which we removed the corresponding images from further steps (anonymizations and training) to isolate privacy risks. In other steps, no such behavior was found, and we used all samples regardless of the caption quality.

\subsection{Diffusion model training and generating}
For \textbf{FreePrompt}~\cite{Liu_2024_CVPR}, following the official implementation, we used an empty string as the source prompt and set the \texttt{SELF\_REPLACE\_STEPS} ratio to 0.4 to encourage substantial appearance changes. Images were generated at a resolution of 512$\times$512 using 50 diffusion steps with a guidance scale of 7.5.

For \textbf{FlowEdit}~\cite{kulikov2025flowedit}, we adopted the SD3 backbone with default hyperparameters provided in the official release. Across all FlowEdit-based experiments, we used $c^{priv}$ as the caption describing the source image.

For \textbf{InstructPix2Pix}~\cite{brooks2023instructpix2pix}, we trained on 4 GPUs with a batch size of 64 and a learning rate of $1\times10^{-5}$. Training was conducted for 200 epochs with gradient accumulation to an effective batch size of 256. We initialized all modules with MagicBrush \cite{zhang2023magicbrush} weights and updated only the UNet parameters. Following the original pipeline, all training images were resized to 256$\times$256, while inference was performed at 512$\times$512 using classifier-free guidance scales of 1.5 for the image and 7.5 for the text prompt, with 100 denoising steps.

We used an identical training and inference setup for the modified \textbf{InstructPix2Pix} equipped with \textbf{Safe Attention}. For initialization, the MagicBrush cross-attention weights were copied into the query, key, value, and output projection layers of the Safe Attention module to ensure compatibility and stable convergence.

For \textbf{OminiControl}~\cite{tan2025ominicontrol}, we adopted the \textit{subject} configuration with a batch size of 4. We used a dummy positional offset of $(0,0)$ to disable spatial displacement and trained for 12{,}000 intervals directly from the FLUX.1 dev~\cite{flux2024} checkpoint. To the best of our knowledge, this makes our work among the first to introduce a privacy-aware, FLUX-based image editing model.

For \textbf{DeepPrivacy2}~\cite{hukkelaas2023deepprivacy2} and \textbf{FaceAnon}~\cite{kung2025face}, we directly applied the default implementations released by the authors to our dataset without modification.

\subsection{Evaluation metrics for Stage~1}
\label{appex:stage1_details}
\paragraph{VISPR attribute grouping.}
VISPR~\cite{orekondy17iccv} provides privacy annotations using attribute identifiers $a_i$. 
In our evaluation, privacy inspection is treated as a binary classification task: 
an image is labeled \emph{safe} if it has attribute \texttt{a0\_safe}, and \emph{unsafe} otherwise.

To analyze performance under specific privacy risks, we further evaluate several attribute groups:

\begin{itemize}
\item \textbf{Face:} \texttt{a9}, \texttt{a10} (complete and partial face)
\item \textbf{Health Indicators:} \texttt{a39}, \texttt{a41}, \texttt{a43} (physical disability, injury, medicine)
\item \textbf{Vehicles:} \texttt{a102}, \texttt{a103}, \texttt{a104} (vehicle ownership, license plate complete/partial)
\item \textbf{Personal Opinion:} \texttt{a61}, \texttt{a62} (general and political opinions)
\end{itemize}

An image is considered privacy-sensitive for a given group if any attribute in the corresponding set is present.

\paragraph{Evaluation protocol.}
For each image, the VLM detector predicts whether privacy-sensitive content is present under the specified criterion. 
Predictions are compared against VISPR annotations, and we report recall, precision, and F1 score. 
Recall is emphasized because missed detections would allow sensitive images to enter the training pipeline. 
All results are computed on the VISPR test split.
\subsection{Evaluation metrics for Stage~2 }
\label{sec:eval_details}

\subsubsection{Cheating Scores}
To measure unintended preservation of perceptual cues from the original image, we compute the Learned Perceptual Image Patch Similarity (LPIPS)~\cite{zhang2018perceptual} with a VGG-16 backbone~\cite{simonyan2014very}. Higher LPIPS indicates stronger deviation from the source image, suggesting that the anonymization process does not simply reconstruct or copy identity-related features.

\subsubsection{Privacy Scores}
Face similarity (\textbf{FaceSim}) is computed using the Antevelop-v2 face encoder~\cite{deepinsight_2023}. For each detected face in the original image, we compute cosine similarity to its nearest counterpart in the anonymized image and retain only the closest match to avoid bias from unmatched pairs.

To evaluate textual leakage (\textbf{TextSim}), we use InternVL2.5-8B~\cite{chen2024expanding} to extract text from anonymized images and compute the token-set similarity using the \texttt{rapidfuzz} package. Lower similarity indicates more effective removal or distortion of identifiable textual content.

For demographic analysis (\textbf{Race Entropy)}, the same VLM predicts demographic attributes (White, Black, Asian, Hispanic, Other), from which we compute the race entropy metric described in Section~\ref{sec:metrics}.

To obtain the \textbf{VLMScore}, we employ InternVL3.5-8B~\cite{wang2025internvl3} as a judge model. The raw and anonymized images are jointly provided to the VLM with a structured prompt asking it to assign a score from 0–100 reflecting how effectively privacy-sensitive attributes have been removed while preserving scene semantics.

\subsubsection{Utility Score}

\paragraph{Classification.}
For classification experiments, we adopt Masked Autoencoders (ImageMAE)~\cite{he2022masked} as the backbone and fine-tune a randomly initialized linear classification head. All models are initialized from ImageNet-pretrained weights. Training uses batch size 64, learning rate $5\times10^{-4}$, gradient accumulation of 4 (effective batch size 256), and 100 epochs. Data augmentation follows the ImageMAE setup using RandAugment with parameters $(n=2, m=9, mstd=0.5)$.

\paragraph{Image Captioning.}
To evaluate semantic preservation in generative tasks, we fine-tune BLIP-2~\cite{blip2} on the filtered MS-COCO dataset using the human-annotated captions. Caption quality is evaluated using BLEU-4 and CIDEr.

\paragraph{Visual Question Answering.}
For VQA, we fine-tune a Qwen3VL-2B~\cite{yang2025qwen3} model on question–answer pairs from the OK-VQA dataset~\cite{okvqa} and report answer accuracy.

In all tasks, model selection is performed on the anonymized validation set constructed in the same manner as the anonymized training set. Final performance is reported on the \textit{original} test sets to measure preservation of task-relevant semantics under the original data distribution. All experiments are conducted on NVIDIA A100 GPUs.

\input{tables/onlysafe}
\input{tables/caption_scores}

\section{Evaluation of Stage~1: Privacy Inspection}

\subsection{Is Image Privacy Anonymization Necessary?} 
\label{appdx:only_safe}
Given the images categorized into safe and unsafe partitions, we need to perform a robust anonymization pipeline on the unsafe images while keep the safe images intact.  We evaluate the extreme setting in which \emph{only} images flagged as safe by the VLM are used for training. This subset inevitably contains some false positives (images incorrectly flagged as safe) which allows us to assess how well the model performs without any anonymization applied to unsafe images.

Table~\ref{tab:onlysafe} summarizes the key statistics. As expected, compared to the model trained on the full (original)  dataset, training solely on the safe subset leads to a substantial drop in downstream utility due to the significant reduction in data volume and diversity. The remaining samples are highly sanitized and lack many of the visual cues needed for effective classification, resulting in noticeably weaker accuracy.

However, the privacy-leakage signals are correspondingly minimal: the VLM detects very few readable texts, faces, or identifiable racial attributes in this safe-only set. This confirms that the privacy detector is conservative and effective as most privacy-sensitive images are successfully excluded. At the same time, the sharp utility degradation highlights the necessity of anonymizing unsafe images rather than discarding them, which helps the model to recover both dataset scale and semantic richness while maintaining strong privacy guarantees.

\subsection{Are Captions Generated by Our Pipeline Helpful for Utility and Quality Preservation?}
\label{appdx:caption_scores}

We provide an evaluation of the captions produced in Stage~1. Although modern VLMs are generally strong captioners, it is important to quantify how well the generated captions align with the underlying visual content, especially when private details are removed or rewritten. To assess caption fidelity, we use the FLEUR benchmark~\cite{fleur}, which measures consistency between the generated caption, the image, and the five human-annotated reference captions from MS-COCO~\cite{lin2014microsoft}. We use the val2014 split, which is also our test set, to obtain the quality score for $c^{priv}$, $c^{pub}$, and $LLM$.
\input{tables/full_results_Omini_FreePrompt}

As shown in Table~\ref{tab:caption_scores}, the public caption $c^{pub}$ exhibits only a modest decrease in FLEUR compared to the private caption $c^{priv}$. This comparable score indicates that the public captions still capture the essential scene semantics that humans perceive while successfully omitting privacy-sensitive content. Notably, the LLM-composed captions also maintain reasonably high alignment despite introducing synthetic, privacy-safe attributes, demonstrating that our edit-instruction generation preserves global scene meaning even when enriching the caption with additional identity-neutral details.

\section{Additional Results of Stage~2: Safe Image Generation}
\label{sec:supp_caption}

\subsection{Quantitative resuls}

In Table~\ref{tab:utility_privacy_full}, we report the performance of our pipeline when leveraging FreePrompt~\cite{Liu_2024_CVPR} and OminiControl~\cite{tan2025ominicontrol} as the editing diffusion models. These results strengthen our conclusion that the \textbf{\sysname} provides a robust framework not only for obtaining privacy-preserving edit instructions, but also for curating an effective dataset for teaching a diffusion model to perform anonymization.

\input{figures/attn_maps}

\subsection{Analysis of attention maps from SafeAttention}
\label{appdx:safe_attn_maps}

 We further analyze the resulting attention maps to visualize how \textbf{SafeAttention} affects the editing behavior. Following Liu et al.~\cite{Liu_2024_CVPR}, we compute the averaged attention maps over all diffusion steps for each of the 16 transformer layers in the UNet. For Safe Cross Attention, we additionally separate the attention maps into the components corresponding to $c^{pub}$ and $c^{edit}$, allowing us to examine how each text source influences the model.

 Figure~\ref{fig:attn_maps} shows the attention maps for the Cross Attention and Safe Cross Attention modules at the 13\textsuperscript{th} transformer layer (counting from 1). As illustrated, in vanilla InstructPix2Pix \cite{brooks2023instructpix2pix}, attention is spread diffusely across the entire image, causing edits to leak into regions that should remain unchanged. In contrast, Safe Cross Attention produces a clear separation when we inspect the maps per token group: public-caption tokens attend primarily to stable background regions and task-relevant objects, while edit-instruction tokens concentrate their attention on the sensitive areas identified for anonymization. This structured attention pattern demonstrates that Safe Cross Attention provides a more controlled and interpretable mechanism for privacy-preserving editing, facilitating denoising that is both more selective and more faithful to the intended anonymization behavior.

\subsection{Does Our Method's Controllability Promote Fairness?}
\label{appdx:controlability}

\input{figures/ethnic}

Throughout the aforementioned experiments, the LLM was free to propose any identity attributes. To assess demographic controllability, we introduced a simple intervention: we constructed a list of racial groups, and for each image, we uniformly sampled one race and asked the LLM to integrate it into the edit ideas. The racial groups include \prompt{White}, \prompt{Black}, \prompt{East Asian}, \prompt{South Asian}, \prompt{Southeast Asian}, \prompt{Middle} \prompt{Eastern/North African}, \prompt{Indigenous/} \prompt{Pacific Islander}, and \prompt{Hispanic/Latino}. \changelogSJcamera{We use demographic predictions as a proxy for diversity.}
Figure~\ref{fig:ethnic} presents an example under the sampled label \prompt{Indigenous/Pacific Islander}.  
After applying our pipeline, the anonymized output reconstructs the scene with entirely new identities whose appearance, such as skin tone, hairstyle, and traditional clothing, aligns with the sampled demographic category. 
At the same time, the global scene geometry and activity are preserved: the individuals are still gathered around the same cake, positioned in the same layout, and engaged in the same collective action. 
The LLM-generated edit instruction also faithfully reflects the sampled demographic, producing a coherent description that guides the editor toward culturally consistent attributes while removing sensitive cues such as uniforms, text, or identifiable faces. This intervention demonstrates that our framework not only anonymizes identity but also provides fine-grained control over demographic attributes simply via text condition when explicitly requested. 

\subsection{Qualitative results}

\input{figures/across_priors}

\input{figures/across_models}
\paragraph{How Are Images Anonymized Differently Across  Text Priors?}  Figure~\ref{fig:across_priors} shows the results when using different text priors to the FlowEdit backbone~\cite{kulikov2025flowedit}. The result for $c^{edit}$ is omitted because the text prior is not suitable for the model, which expects a description of the target scene rather than edit instructions.

We observe that existing face anonymization frameworks fail to properly address non-human privacy concerns and instead prioritize modifying the entire face or body, which can sometimes be missed. In contrast, our method applies editing to the whole image and preserves only the information necessary for downstream tasks.

It is important to note that the priors $c^{priv}$ and $c^{class}$ represent unprocessed information that is not fully derived from our Stage 1 and is typically available in common image datasets, such as human-annotated captions or class labels. These priors are not recommended in our pipeline and tend to either fail to protect sensitive attributes or discard valuable diversity in the images.
\input{figures/scatterplots}

%Analyze the museum example
In comparison, safe text prompts including $c^{pub}$, $c^{edit}$, and ${LLM}$ are much more suitable for effective downstream learning. This improvement is due to their alignment with the original content as well as their ability to preserve fine-grained details and structural dynamics.

\paragraph{How Are Images Anonymized Differently Across  Generative Backbones?} We also show the quality of anonymization across different backbone diffusion models. Figure~\ref{fig:across_models} shows outputs of all 4 diffusion models with the most compatible text prior as suggested by their authors: FreePrompt with $c^{edit}$, finetuned InstructPix2Pix with $c^{edit}$, FLUX-based OminiControl with $c^{edit}$, and FlowEdit with the target caption ${LLM}$. Regardless of the generator choice, our \sysname generates images that highly align to the original content while revealing little leakage.

\section{Visualized Trade-Off Among All Evaluation Dimensions}

Figure~\ref{fig:scatter} summarizes the quantitative evaluation across multiple dimensions. 
Each subplot uses a different \textit{privacy} indicator on the x-axis (Face Similarity, Text Similarity, or Race Entropy) and downstream \textit{utility} (top-1 accuracy) on the y-axis. 
Marker size reflects the \textit{cheating} structural similarity (SSIM), while color, shape, and outline respectively encode the generative backbone, the text prior used during editing, and whether the model was fine-tuned.

This consolidated visualization makes the privacy-utility-fidelity trade-off directly interpretable.  Face-only anonymization baselines perform well on face similarity and achieve slightly higher classification accuracy, but they fail to remove low-level cues, textual information, and demographic signals, resulting in substantially weaker overall privacy protection. In contrast, as revealed by the scatterplots, our editing models generally occupy \textit{favorable} regions of the trade-off space: they reduce identifiable content while maintaining competitive utility and without relying on structural similarity.

\section{Prompts to Language Models}
\label{sec:supp_promts}
\subsection{Prompts used in Stage 1}
\label{sec:supp_promts_stage1}

Figures~\ref{fig:prompt_criteria}, \ref{fig:prompt_flag}, \ref{fig:prompt_caption}, \ref{fig:prompt_edit}, and \ref{fig:prompt_combined} illustrate the prompting components that shape how Stage 1 decomposes privacy reasoning into a sequence of explicit and controllable text-generation steps.
Figure~\ref{fig:prompt_criteria} defines the privacy criteria used throughout Stage 1 by summarizing VISPR~\cite{orekondy17iccv}’s 67 attributes into nine interpretable categories.

Figure~\ref{fig:prompt_flag} shows the prompt used to obtain the \texttt{PRIVACY\_FLAG}. This prompt is intentionally conservative, defaulting to \texttt{TRUE} under any ambiguity, to minimize false negatives, since any missed detection would allow sensitive content to pass downstream. The expected output is intentionally short and “explanation-free,” facilitating fast, large-scale screening across all images.

Figure~\ref{fig:prompt_caption} presents the structured captioning prompt, which serves as the backbone of Stage 1. Although the \texttt{PRIVACY\_REVIEW} is not directly used in later steps, it forces the VLM to explicitly enumerate privacy-relevant elements, thereby making omissions auditable and ensuring that the subsequent private and public captions are grounded in a clear semantic separation.

Figure~\ref{fig:prompt_edit} provides the prompt used by the LLM to generate edit instructions. It requires attribute-level rewriting (e.g., gender, hair, clothing, body shape, cultural markers) while prohibiting repetition of unchanged details. Importantly, the image itself is not provided to the LLM at this stage; the model must reason solely from the privacy-preserving caption.

Finally, Figure~\ref{fig:prompt_combined} shows the prompt used to merge the public caption and the edit instruction into a single compact description. Despite its simplicity, this step addresses two practical constraints: (i) diffusion editors often impose strict token budgets, and (ii) the merged caption enables seamless integration of available ground-truth labels for preserving task-relevant semantics during editing.

\input{text_blocks/prompt_criteria}
\input{text_blocks/prompt_flagging}
\input{text_blocks/prompt_captioning}
\input{text_blocks/prompt_edit_instruction}
\input{text_blocks/prompt_combine_llm}

\label{sec:supp_promts_eval}

\input{text_blocks/prompt_custom_criteria}
\input{text_blocks/prompt_ocr}
\input{text_blocks/prompt_demographic}
\input{text_blocks/prompt_vlm_compare}
\subsection{Prompts used in evaluation}
We report here the prompts used for all VLM-based evaluation components. These cover four tasks:  
(1) privacy flagging under a restricted set of criteria,  
(2) extraction of readable text from the image,   
(3) detection of demographic attributes, and
(4) judement of anonymization quality.  
The corresponding prompts are shown in Figure~\ref{fig:prompt_custom_criteria}, Figure~\ref{fig:prompt_ocr},  Figure~\ref{fig:prompt_demographic}, and Figure~\ref{fig:prompt_judge}, respectively.

% For the demographic inspection task, the VLM was queried about both race and gender. However, in our quantitative analysis we focus exclusively on \emph{race} distributions. Gender predictions were deliberately excluded for two reasons. First, gender presentation is highly ambiguous in many anonymized outputs and cannot be reliably inferred without risking misclassification or introducing unintended biases. Second, our primary goal is to measure demographic \emph{uniformity} and \emph{diversity} across anonymized datasets, for which race-based entropy provides a more stable and interpretable signal. Accordingly, race uniformity is the only demographic metric incorporated into our evaluation framework.

%\input{figures/gradioUI}
% \clearpage
\section{Deployment}
We provide the implementation for a user-friendly Graphical User Interface (GUI) of our pipeline via Gradio \cite{abid2019gradio}. The interface demonstrates how an uploaded image is automatically processed by our VLM- and LLM-based inspectors to produce safe captions and an anonymized proxy generated by the selected editing model.

To promote accessibility and reproducibility, we plan to publicly release this interface to the research community. 
%after the review process.

% To support reproducibility while maintaining anonymity, we host the interface through an anonymous temporary Gradio link, allowing reviewers to test the pipeline without violating double-blind review requirements.

%% file: figures/clip_score_distr.tex
\begin{figure}[t]
\centering
% \vspace{6em}
%\fbox{\rule{0pt}{2in} \rule{.95\linewidth}{0pt}}
\resizebox{0.9\columnwidth}{!}{\includegraphics[]{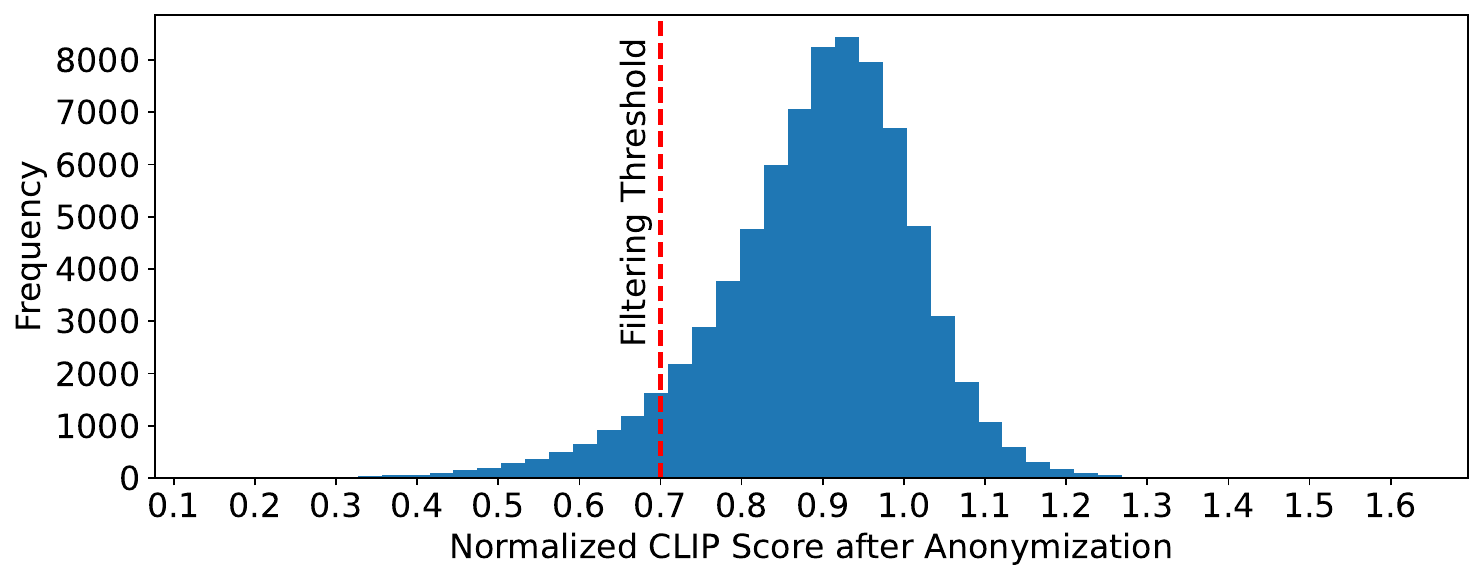}}
\caption{
\textbf{Distribution of normalized CLIP scores for private images after anonymization.} 
Most samples cluster around 0.9, indicating strong preservation of original semantics. 
The red line marks the filtering threshold at 0.7, below which edited samples are discarded during editing model training.
}

\label{fig:clip_score_distr}
\end{figure}

%% file: tables/onlysafe.tex
\begin{table}[t]
\centering
\caption{Summary of dataset statistics, utility accuracy, and privacy-leakage indicators for the raw and safe subsets. The table reports training/validation sample counts, top-1 accuracy, and the number of detected text, faces, and race attributes.}
\resizebox{\columnwidth}{!}{%
\begin{tabular}{lcccccc}
\toprule
& \multicolumn{2}{c}{\textbf{\#Samples}} 
& \multicolumn{1}{c}{\textbf{Utility}} 
& \multicolumn{3}{c}{\textbf{Privacy Instances}} \\
\cmidrule(lr){2-3} \cmidrule(lr){4-4} \cmidrule(lr){5-7}
\textbf{Split} 
& \textbf{Train} 
& \textbf{Val} 
& \textbf{Acc@1} 
& \textbf{Text} 
& \textbf{Faces} 
& \textbf{Races} \\
\midrule
\multicolumn{7}{l}{\textbf{Caltech101}~\cite{caltech101}} \\
\; Original & 2000 & 980 & 94.27  & 251 & 82  & 71 \\
\; Safe subset & 1552 & 756 & 83.49&  111 & 24  & 0  \\
\midrule
\multicolumn{7}{l}{\textbf{MIT Indoor67}~\cite{MITindoor67}} \\
\; Original & 3991 & 1317 & 83.88 & 551 & 2184 & 1332 \\
\; Safe subset & 1605 & 512  & 51.12&  84  & 78   & 1    \\

\bottomrule
\end{tabular}}
\label{tab:onlysafe}
\end{table}

%% file: tables/caption_scores.tex
\begin{table}[t]
\centering
\small
\caption{{\textbf{Alignment to MS-COCO annotations of different text priors produced by Stage~1 under the FLEUR \cite{fleur} metric} (higher is better). The public caption $c^{pub}$ maintains FLEUR scores close to the private caption $c^{priv}$, indicating strong preservation of global scene semantics despite removing sensitive details. The $LLM$ prior , while lower due to the introduction of additional synthesized attributes, still retains meaningful semantic alignment. }}

{%
\begin{tabular}{c|ccc}
\toprule
\textbf{Text Prior} & $c^{priv}$ & $c^{pub}$ & $LLM$ \\
\midrule
\textbf{FLEUR ($\uparrow$)}  & 80.45      & 78.93     & 58.27       \\
\bottomrule
\end{tabular}}
\label{tab:caption_scores}
\end{table}

%% file: tables/full_results_Omini_FreePrompt.tex
\begin{table*}[h]
\centering
\footnotesize
\setlength{\tabcolsep}{6pt}
\renewcommand{\arraystretch}{1.1}
\caption{
\textbf{Extend utility and privacy comparison for different generative backbones.} 
Our \sysname, which uses \cite{kulikov2025flowedit} as the underlying image editor, 
achieves comparable performance on downstream tasks while successfully anonymizing 
privacy-sensitive information, unlike other anonymization models. 
The \textbf{best} value (yellow) and the \textbf{second-best} value (blue) are highlighted per column. $^\dagger$ denotes the model was finetuned with our dataset.
}
\begin{tabular}{ll|cc|cc|cc|cc}
\toprule
\multirow{3}{*}{\textbf{Model}} &
\multirow{3}{*}{\textbf{Text Prior}} &
\multicolumn{2}{c|}{\textbf{Utility Score}} &
\multicolumn{6}{c}{\textbf{Privacy Score}} \\
\cmidrule(lr){3-4}
\cmidrule(lr){5-10}
& &
\multicolumn{2}{c|}{\textbf{Accuracy ($\uparrow$)}} &
\multicolumn{2}{c|}{\textbf{FaceSim ($\downarrow$)}} &
\multicolumn{2}{c|}{\textbf{TextSim ($\downarrow$)}} &
\multicolumn{2}{c}{\textbf{Race Entropy ($\uparrow$)}} \\
& &
Cal101 & Indoor &
Cal101 & Indoor &
Cal101 & Indoor &
Cal101 & Indoor \\
\midrule

% ---------------- Raw images ----------------
Raw Images & -- &
94.277 & 83.881 &
1.0000 & 1.0000 &
1.0000 & 1.0000 &
0.4384 & 0.7443 \\
\midrule

% ---------------- FlowEdit block ----------------
\multirow{5}{*}{\textbf{\sysname} (FlowEdit~\cite{kulikov2025flowedit})}
 & $c_{\text{private}}$ &
94.334 & 79.925 &
0.4378 & 0.2666 &
0.6611 & 0.4210 &
{0.5552} & 0.7399 \\

 & class &
93.857 & 80.448 &
0.4965 & 0.3743 &
\cellcolor{bestcell}{\textbf{0.4856}} & \cellcolor{bestcell}{\textbf{0.2077}} &
{0.4051} & 0.6508 \\

 & $c_{\text{public}}$ &
94.487 & 80.746 &
0.4881 & 0.2883 &
0.5395 & 0.2896 &
{0.6409} & 0.7208 \\

 & $c_{\text{edit}}$ &
{94.792} & 77.090 &
0.3658 & \cellcolor{secondcell}{\textbf{0.2077}} &
0.5238 & 0.2393 &
0.7646 & 0.7589 \\

 & LLM ($c_{\text{edit}},c_{\text{public}}$) &
92.884 & 80.746 &
\cellcolor{bestcell}{\textbf{0.3428}} & 0.2294 &
\cellcolor{secondcell}{\textbf{0.4881}} & \cellcolor{secondcell}{\textbf{0.2119}} &
\cellcolor{bestcell}{\textbf{0.8751}} & {0.7643}\\
\midrule
% ---------------- FreePrompt block ----------------
\multirow{5}{*}{\textbf{\sysname} (FreePrompt~\cite{Liu_2024_CVPR})}
 & $c_{\text{private}}$ &
\cellcolor{bestcell}{\textbf{94.926}} & 81.567 &
0.5026 & 0.2764 &
0.6382 & 0.2757 &
0.5421 & 0.6963 \\

 & class &
94.506 & 79.254 &
0.5474 & 0.2966 &
0.5651 & 0.2314 &
0.4991 & 0.5960 \\

 & $c_{\text{public}}$ &
\cellcolor{secondcell}{\textbf{94.849} }& 82.537 &
0.5085 & 0.2811 &
0.5949 & 0.2423 &
0.5790 & 0.6899 \\

 & $c_{\text{edit}}$ &
93.857 & 78.881 &
0.3693 & 0.2165 &
0.5672 & 0.2361 &
0.7516 & \cellcolor{bestcell}{\textbf{0.8276}} \\

 & LLM ($c_{\text{edit}},c_{\text{public}}$) &
94.105 & 80.522 &
0.4539 & 0.2159 &
0.5580 & 0.2217 &
{0.7967} & 0.7806 \\
\midrule
% ---------------- Omini block ----------------
\multirow{1}{*}{\textbf{\sysname} (OminiControl~\cite{tan2025ominicontrol})$^\dagger$}
& $c_{\text{edit}}$ & 94.506 &	80.000 &	\cellcolor{secondcell}{\textbf{0.3585}} &	\cellcolor{bestcell}{\textbf{0.1925}} &	0.6132 &	0.3350 &	\cellcolor{secondcell}{\textbf{0.8425}} &	\cellcolor{secondcell}{\textbf{0.8058}}\\
\midrule
% ---------------- Baselines ----------------
DeepPrivacy2~\cite{hukkelaas2023deepprivacy2} & -- &
94.601 & \cellcolor{bestcell}{\textbf{84.030}} &
0.3921 & 0.3547 &
0.9569 & 0.8653 &
0.7315 & 0.7547 \\

FaceAnonSimple~\cite{kung2025face} & -- &
\cellcolor{secondcell}{\textbf{94.849}} & \cellcolor{bestcell}{\textbf{84.030}} &
0.4586 & 0.5045 &
0.9355 & 0.7701 &
0.6091 & 0.7407 \\

\bottomrule
\end{tabular}

\label{tab:utility_privacy_full}
\end{table*}

%% file: figures/attn_maps.tex
\begin{figure*}[h]
\centering

\resizebox{0.9\linewidth}{!}{%
    \includegraphics[trim=0pt 400pt 0pt 290pt, clip]{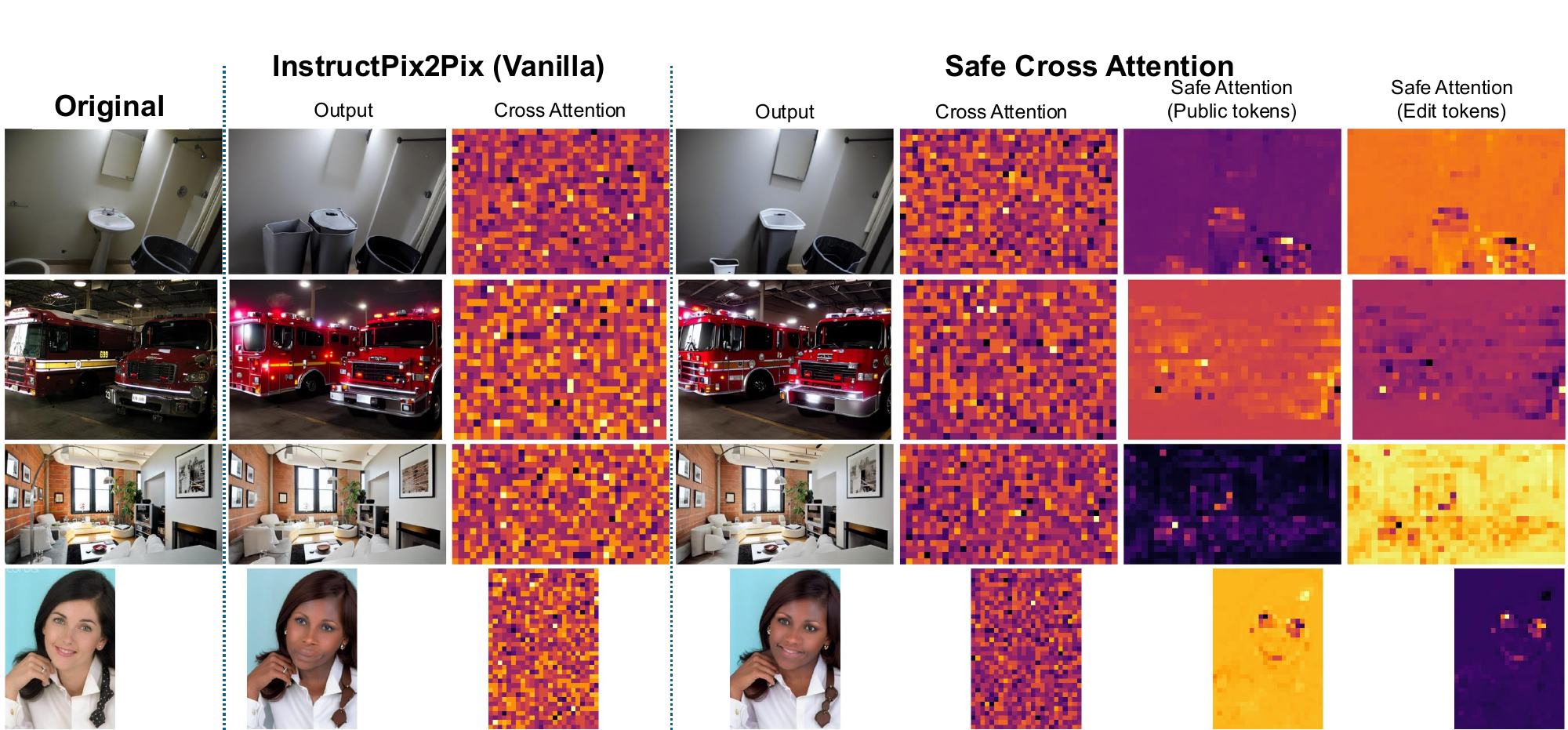}
}

\caption{
Attention maps at the 13\textsuperscript{th} transformer block in the UNet. From left to right:
(1) the original private image;
(2) the anonymized output produced by vanilla InstructPix2Pix along with its standard cross-attention map; (3) the anonymized output produced by our Safe Cross Attention model, shown with its vanilla cross-attention map, the Safe Attention map corresponding to public-caption tokens, and the Safe Attention map corresponding to edit-instruction tokens.
}
\label{fig:attn_maps}
\end{figure*}

%% file: figures/ethnic.tex
\begin{figure}[th]
\centering

\resizebox{\linewidth}{!}{%
    \includegraphics[trim=400pt 100pt 500pt 150pt, clip]{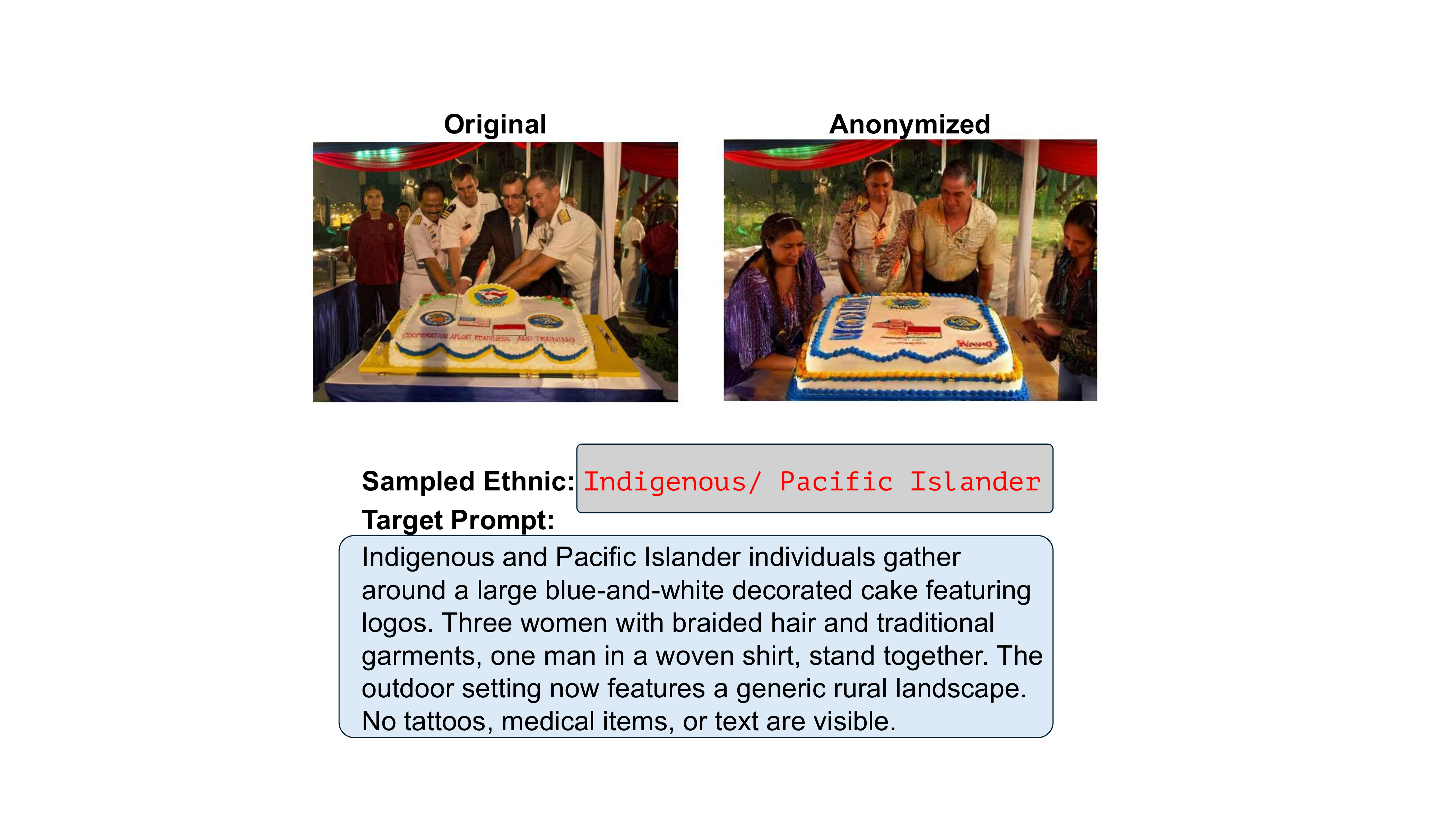}
}

\caption{
 Example of demographic-controlled anonymization under the \textit{randomly} sampled “Indigenous / Pacific Islander” condition.
}
\label{fig:ethnic}
\end{figure}

%% file: figures/across_priors.tex
\begin{figure*}[]
\centering
% \vspace{-20pt}
\resizebox{0.85\linewidth}{!}{%
    \includegraphics[trim=0pt 600pt 0pt 0pt, clip]{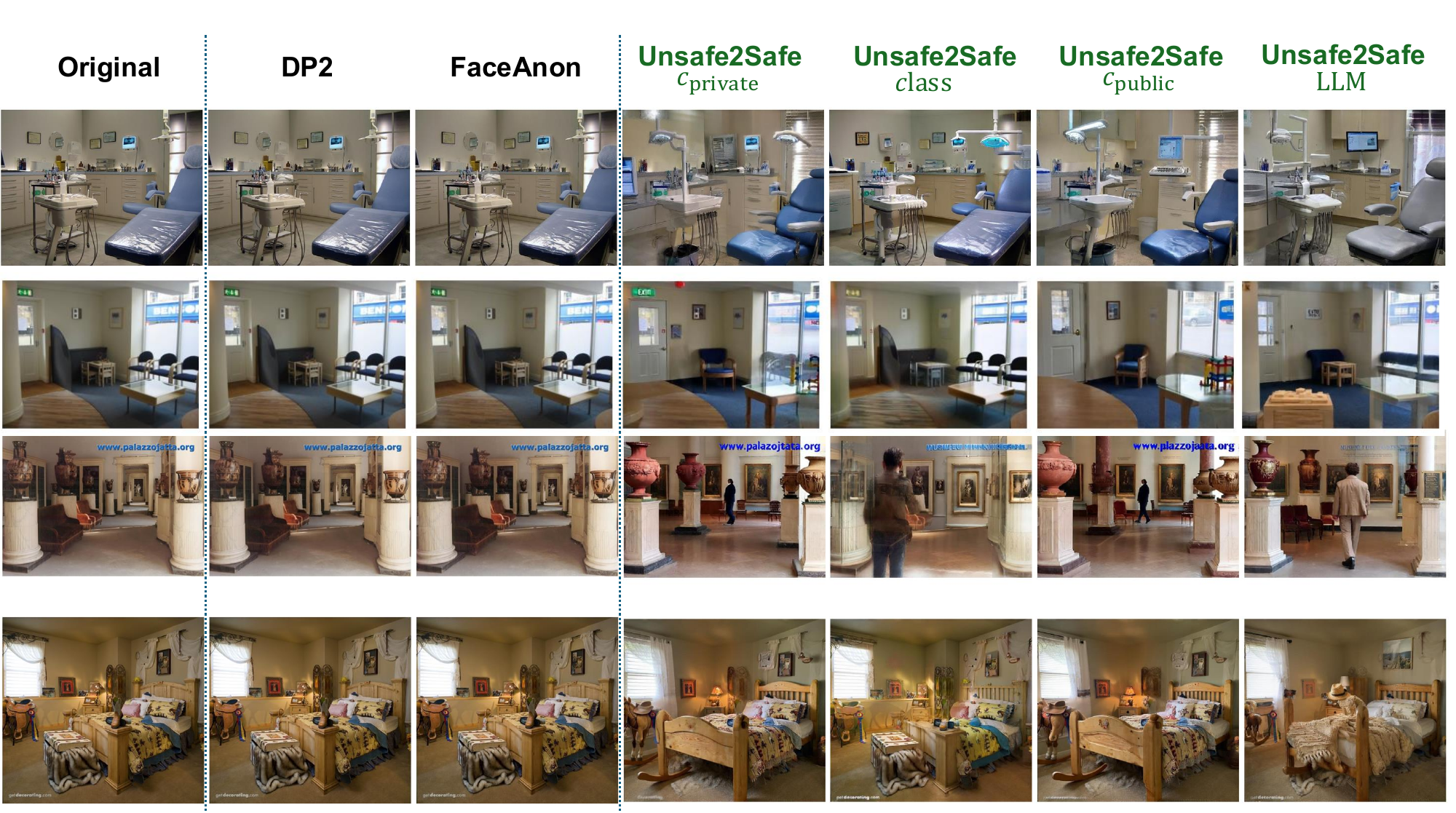}
}

\caption{
Qualitative results of our method with different text priors and existing face anonymizers. Our results were generated by the FlowEdit~\cite{kulikov2025flowedit} model.  
}
\label{fig:across_priors}
\end{figure*}

%% file: figures/across_models.tex
\begin{figure*}[t]
\centering

\resizebox{0.8\linewidth}{!}{%
    \includegraphics[trim=0pt 460pt 0pt 35pt, clip]{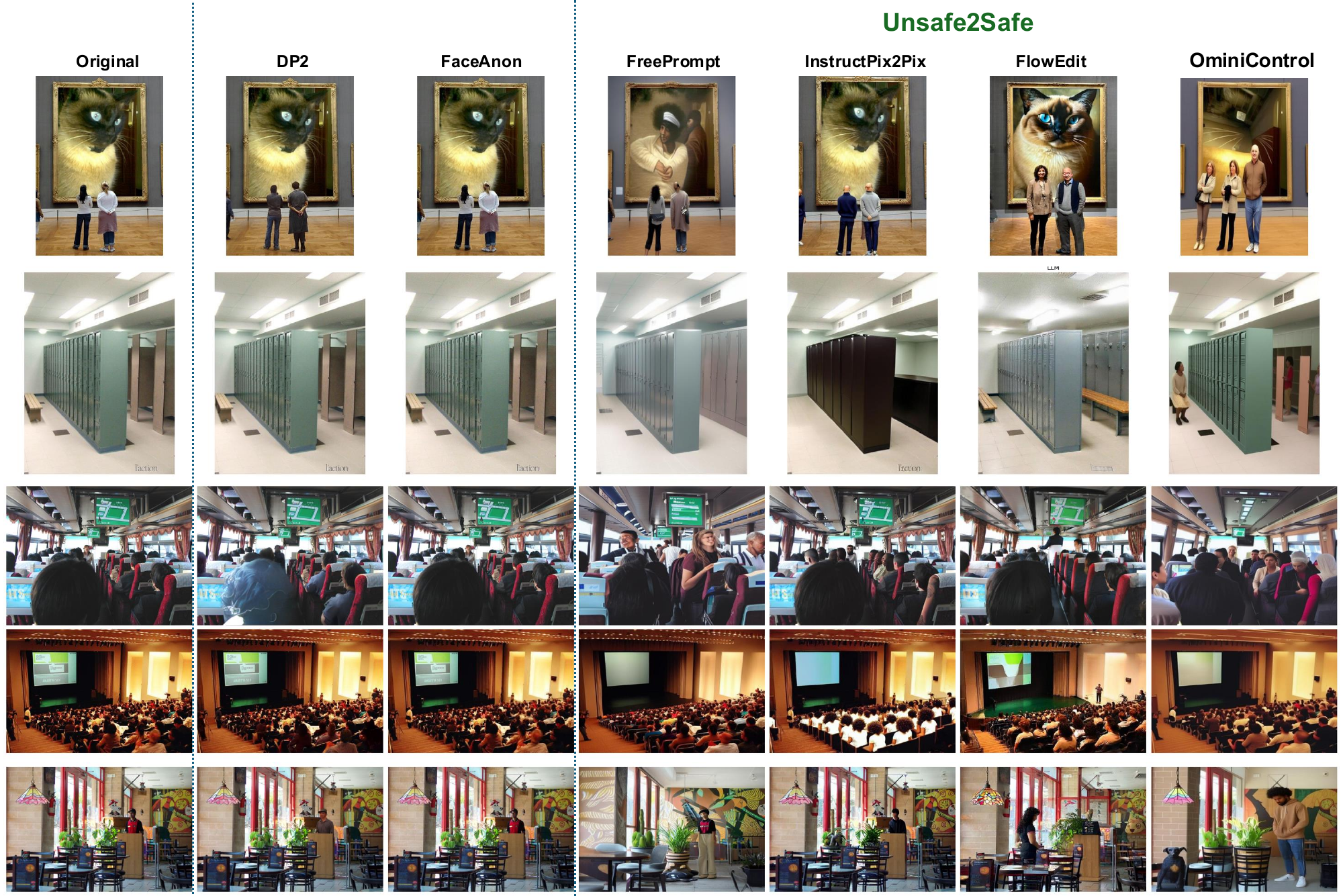}
}

\caption{
Qualitative results of our method with different generative backbones and existing face anonymizers. FlowEdit~\cite{kulikov2025flowedit} takes $LLM$ as the target text prior, while other models take $c^{edit}$ as the text prior.  
}
\label{fig:across_models}
\end{figure*}

%% file: figures/scatterplots.tex
\begin{figure*}[t]
\centering

\resizebox{0.9\linewidth}{!}{%
    \includegraphics[trim=0pt 950pt 0pt 720pt, clip]{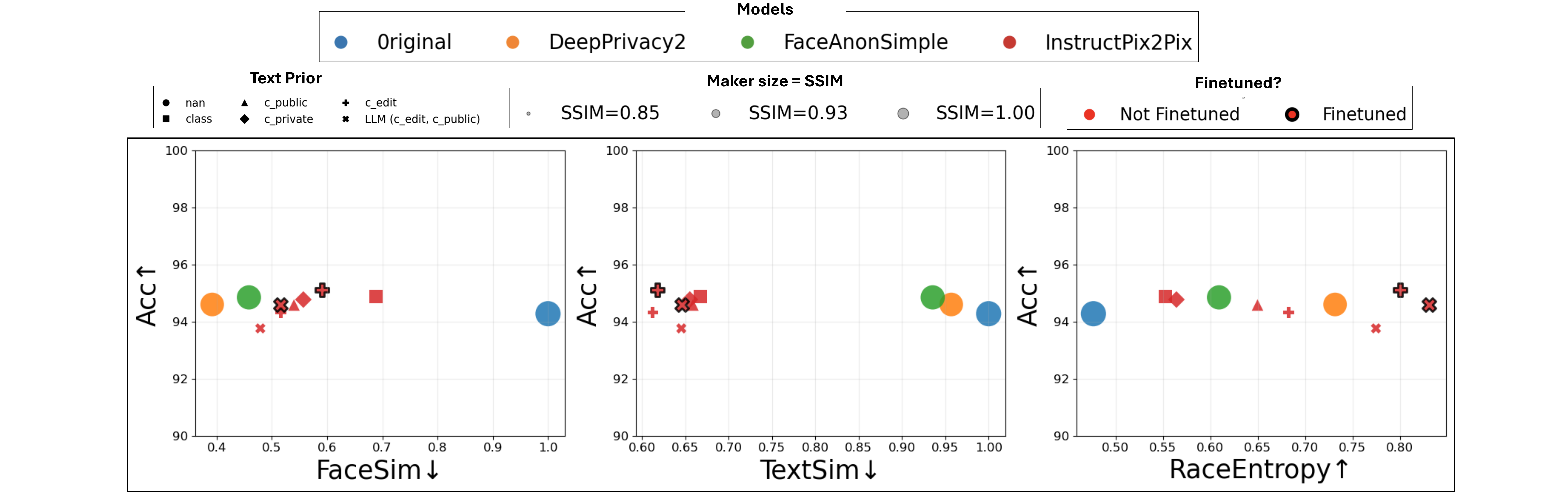}
}

\caption{
Multi-dimensional evaluation of privacy--utility trade-offs across anonymization models with the Caltech101 dataset~\cite{caltech101}. 
Each subplot visualizes one privacy metric (FaceSim, TextSim, or RaceEntropy) against downstream utility (Acc@1). 
Colors denote different editing backbones, and marker shapes represent the text priors used during generation. 
Marker size is proportional to the SSIM \textit{cheating} score, and a black outline indicates models fine-tuned on our anonymization dataset. 
This visualization highlights how fine-tuning and text conditioning jointly influence privacy protection, realism, and task performance.
}
\label{fig:scatter}
\end{figure*}

%% file: text_blocks/prompt_criteria.tex
\begin{figure*}[h!]
    \centering
    % \paragraph{Prompt for Caption Generation}
    \begin{Verbatim}[fontsize=\scriptsize,commandchars=\\\{\}]
- **People**, faces (full or partial), visible skin in private contexts, tattoos, or emotional expressions  
- **Appearance cues**: clothing, accessories, or features revealing gender, culture, religion, status, or roles  
- **Health indicators**: casts, wheelchairs, bandages, medications, disabilities, or visible illness/injury  
- **Texts or symbols**: signs, receipts, documents, screens, handwriting, license plates, usernames, phone numbers  
- **Locations or timestamps**: homes, offices, bedrooms, addresses, classrooms, GPS data, clocks, date/time stamps  
- **Personal interactions**: small groups, families, friends, teams, protests, celebrations, or close relationships  
- **Professional/school settings**: uniforms, tools, name tags, diplomas, workspaces, formal events  
- **Cultural/political/religious identifiers**: flags, slogans, attire, ceremonies, posters  
- **Private belongings or behaviors**: hobbies, photos, mail, social media, travel documents, receipts
 \end{Verbatim}
   \caption{
Privacy-related visual criteria used in the Stage 1 of our pipeline. These criteria are summarized from the set of 67 criteria introduced by VISPR \cite{orekondy17iccv}.
}
\label{fig:prompt_criteria}
\end{figure*}

%% file: text_blocks/prompt_flagging.tex
\begin{figure*}[h!]
    \centering
    % \paragraph{Prompt for Caption Generation}
    
    \begin{Verbatim}[fontsize=\scriptsize,commandchars=\\\{\}]
SECTION: PRIVACY_FLAG
Determine whether the image contains **any content that could disclose private, sensitive, or 
personally identifying information**. Do not include explanations, justification, or extra text.
---
Set: **PRIVACY_FLAG: TRUE**  
if the image contains *even partially visible*, *blurred*, or *contextually implied* content from 
the list below. If uncertain, ambiguous, low-resolution, or partially occluded—**default to TRUE**.
---
Flag the image if you observe or reasonably infer the presence of:
[CRITERIA]
---
**Treat these ambiguous cases as DEFINITE PRIVACY_FLAG: TRUE**:  
- Partial faces in mirrors or reflections  
- Blurred or occluded name tags, signs, or screens  
- Visible groups of people even without faces  
- Partially legible receipts or handwritten notes  
- Niche personal objects (e.g. medals, family photos, memorabilia)
 \end{Verbatim}
   \caption{Assistant prompt used for the privacy filtering step in \textbf{Stage 1}. When a private issue in any image is detected, the VLM sets \texttt{PRIVACY\_FLAG} to \texttt{TRUE}, even when the issues were partially visible, contextually implied, ambiguous, or low-resolution cases.}

\label{fig:prompt_flag}
\end{figure*}

%% file: text_blocks/prompt_captioning.tex
\begin{figure*}[!h]
    \centering
    % \paragraph{Prompt for Caption Generation}
    
    \begin{Verbatim}[fontsize=\scriptsize,breaklines=true,commandchars=\\\{\}]
Your task is to analyze the image and generate **structured, privacy-aware captions** in 3 sections:  
- `PRIVACY_REVIEW`: Identify sensitive or private elements 
- `PRIVATE_CAPTION`: Full factual visual description  
- `PUBLIC_CAPTION`: Safe, anonymized caption suitable for generative training
---
SECTION: PRIVACY_REVIEW
List up to 10 elements in the image that pose a privacy risk. For each, describe what is visible 
and why it is sensitive. Use this format:
- ITEM: [Short factual phrase of the visible element]  
- REASON: [Why it is privacy-sensitive, based on the list below]
Consider the following privacy risk categories:
[CRITERIA]
Example:
- ITEM: A woman hugging a child in a decorated living room 
- REASON: Relationship, identity, and private setting
---
SECTION: PRIVATE_CAPTION
Provide a richly detailed narrative (<= 50 tokens) of everything visible in the image, even if 
privacy-sensitive. Include:
- People, animals, objects, or vehicles and their appearances
- Visible logos, brand names, or text
- The full environment and any identifying or contextual features
---
SECTION: PUBLIC_CAPTION
Write a **safe and fluent caption** (<= 50 tokens) that:
- Rephrases or removes all sensitive elements listed in PRIVACY_REVIEW.
- Do NOT include identity, brand names, text, or logos
- Focuses on scenery, layout, lighting, objects, and anonymous presence (e.g., “a person” is allowed 
if not identifiable)  
- Use fluent, image-grounded language. Do **not** censor awkwardly or produce incomplete fragments. 
The caption should stand alone.
 \end{Verbatim}
   \caption{
User prompts provided to the VLM for generating structured, privacy-aware captions of private images. The prompt consists of three sections: \texttt{PRIVACY\_REVIEW}, which identifies potentially sensitive elements; \texttt{PRIVATE\_CAPTION}, which offers a detailed caption of the entire scene; and \texttt{PUBLIC\_CAPTION}, which produces a concise, privacy-compliant caption suitable for use as guidance for diffusion models.
}
\label{fig:prompt_caption}
\end{figure*}

%% file: text_blocks/prompt_edit_instruction.tex
\begin{figure*}[!h]
    \centering
    % \paragraph{Prompt for Caption Generation}
    
    \begin{Verbatim}[fontsize=\scriptsize,breaklines=true,commandchars=\\\{\}]
You are given a caption that describes a real-world photo:
Caption: {public_caption}
Your task is to propose realistic identity modifications while the content of the scene (layout, objects, 
actions) must remain exactly the same.
These descriptions will be applied as editing instructions for a Stable Diffusion model.
- Make the modifications significant, specific, and easy to generate by a Stable Diffusion.
- Change every person and every object with privacy risks.
- For each person, include at least 3 specific changes regarding gender, hair, body shape, skin 
tone, clothing, or culture.
- Neutralize objects or background cues with privacy risks with plain or generic alternatives.
The changes may include:
[CRITERIA]
Rules:
- Preserve the current object appearence, layout, or actions.
- Output only the modifications, not rewrite of the caption.
- Do NOT repeat the unchanged details from the original caption. 
- Keep it realistic, specific, and <= 45 words continuous, natural phrase fragment
 \end{Verbatim}
\caption{ Prompt to the  LLM for generating  edit ideas as instructions conditioned on the public caption. The model proposes identity-neutral but realistic attribute modifications, with a focus on faces, while preserving layout and scene semantics. 
}
\label{fig:prompt_edit}
\end{figure*}

%% file: text_blocks/prompt_combine_llm.tex
\begin{figure}[!h]
    \centering
    % \paragraph{Prompt for Caption Generation}
    
    \begin{Verbatim}[fontsize=\footnotesize,breaklines=true,commandchars=\\\{\}]
You are given a caption describing a real-world 
photo and an edit instruction. 
Original caption: {public_caption}
Edit instruction: {edit_caption}

Combine them into a single, natural-sounding caption 
that reflects ALL edits while preserving the 
unchanged parts. Keep the result concise (under 
50 words). After combining, the caption has to
describe an image of {class_name}.
 \end{Verbatim}
   \caption{Prompt for merging the public caption and edit instruction into a single compact description compatible with token-limited diffusion editors. When a class name or ground-truth annotation is available, users can simply include it into the prompt for a task-relevant caption.
}
\label{fig:prompt_combined}
\end{figure}

%% file: text_blocks/prompt_custom_criteria.tex
\begin{figure}[]
    \centering
    % \paragraph{Prompt for Caption Generation}
    
\begin{Verbatim}[fontsize=\scriptsize,commandchars=\\\{\}]
SECTION: PRIVACY_FLAG
Determine whether the image contains any instance 
of **{criteria}**. Respond with only one line:
**PRIVACY_FLAG: TRUE** or **PRIVACY_FLAG: FALSE**
Do not include explanations or additional text.
---
**Set PRIVACY_FLAG: TRUE** if the image clearly, 
partially, or contextually depicts any CRITERIA_NAME. 
If the presence of {criteria} is uncertain, ambiguous, 
low-resolution, or partially occluded, **default to 
TRUE**.
**Set PRIVACY_FLAG: FALSE** if the image does not 
contain {criteria}, or if any privacy concern is 
unrelated to {criteria}.
---
Flag the image as TRUE if you can observe or 
reasonably infer any of the following:
CRITERIA_DESCRIPTION

 \end{Verbatim}
   \caption{
Prompt used to ablate the flagging step with custom, less-inclusive privacy criteria.
}
\label{fig:prompt_custom_criteria}
\end{figure}

%% file: text_blocks/prompt_ocr.tex
\begin{figure}[!h]
    \centering
    % \paragraph{Prompt for Caption Generation}
    
    \begin{Verbatim}[fontsize=\scriptsize,commandchars=\\\{\}]
You are an image analysis assistant. Analyze the 
provided image and extract all visible text that 
appears anywhere in the scene (including logos, 
signs,  menus, or labels). If no readable text is 
visible, output exactly: NO_TEXT

Otherwise, respond strictly in the following format:
### TEXT
<verbatim text as it appears, preserving 
capitalization, punctuation, and line breaks>

Rules:
- Include only the text visible in the image - no
descriptions or interpretations.
- Preserve exact wording, capitalization, 
and line breaks.
- If part of the text is unclear, replace 
it with [unclear].
- Output exactly one block following the format 
above.
- If no text is visible, output only "NO_TEXT".
\end{Verbatim}
   \caption{User prompt used for detecting text in the image.}

\label{fig:prompt_ocr}
\end{figure}

%% file: text_blocks/prompt_demographic.tex
\begin{figure}[!h]
    \centering
    % \paragraph{Prompt for Caption Generation}
    
    \begin{Verbatim}[fontsize=\scriptsize,commandchars=\\\{\}]
You are an image analysis assistant. Analyze the 
provided image and identify all apparent genders 
and races  visible in the scene. If no clearly 
visible person (no recognizable face or body) 
is present, output exactly: NO_HUMAN
Otherwise, respond strictly in the following format:
### GENDER
<male and/or female>
### RACE
<list of races: White, Black, Asian, Hispanic, 
Middle Eastern, Indian, Other>
Rules:
- Use only the categories provided.
- Include all that clearly appear, separated by commas.
- Do NOT include "unknown" or any explanation.
- Output exactly one block following the format above.
- If no human is visible, output only "NO_HUMAN".
\end{Verbatim}
   \caption{User prompt used for detecting demographical attributes (gender and race) in the image.}

\label{fig:prompt_demographic}
\end{figure}

%% file: text_blocks/prompt_vlm_compare.tex
\begin{figure*}[!h]
    \centering
    % \paragraph{Prompt for Caption Generation}
    
    \begin{Verbatim}[fontsize=\footnotesize,breaklines=true,commandchars=\\\{\}]
System Prompt: You are a vision-language evaluator for privacy-preserving image anonymization. You must compare the left and right images and judge anonymization strictly based on visual differences. Assign a score from 0 to 100 based on how much that difference reduces the ability to identify or link specific people, objects, or attributesfrom the LEFT image to the RIGHT image. Do not infer or assume sensitive information beyond what is visually identifiable.

User Prompt:
You are given two images shown side by side:
- Left image: the original image
- Right image: the edited / anonymized image
Your task is to assign an anonymization score that measures how much the RIGHT image reduces the ability to link the LEFT image to the same specific identities or personal specifics.
Privacy-sensitive information includes (but is not limited to):
[CRITERIA]
You MUST follow this decision process:
STEP 1 — Instance identity check:
Decide whether the LEFT and RIGHT images depict the same characterstics of privacy-senstive information.
If all of them they appear to be the same instance with no meaningful visual change, the score will be 0.
- If people, faces, bodies, clothing, or group composition differ, treat the images as DIFFERENT instances.
- The same layout, camera angle, or furniture arrangement does NOT imply the same instance.
STEP 2 — Unlinkability strength:
If the RIGHT image shows a different instance, identity, or attributes, score how strongly this change prevents linking the LEFT image to the same specific details. The general scene, activity, or layout may remain similar.
Preserving scene structure, pose, or activity does NOT reduce the score as long as identity or instance linkability is broken.
---
### Scoring (0–100)
- **0**: Same identities or same instance (no anonymization).
- **20–40**: Weak unlinkability (small but real instance or attribute change).
- **40–70**: Moderate unlinkability (multiple identity-related cues changed).
- **70–90**: Strong unlinkability (identities or specifics are hard to link).
- **90–100**: Near-complete unlinkability of identities or personal specifics.
High scores are allowed even when scene/layout/activity is preserved, if identities or spersonal specifics are replaced or neutralized.
Reward evidence of:
- people/face identity replacement or strong masking (blur/occlusion/distortion)
- removal/replacement of tattoos, distinctive clothing/accessories, name tags
- removal/replacement of readable text (documents, screens, plates, usernames)
- replacement of distinctive personal objects or environment instance details
---
Rules:
- Give score 0 for identical pair.
- Give score 100 for totally different image pair.
- Do NOT penalize preservation of scene structure or activity.
- Ignore image quality, realism, or aesthetics.
- Do not infer sensitive information that is not clearly visible.
Output:
ANONYMIZATION_SCORE: <integer 0-100>
 \end{Verbatim}
\caption{Prompt to the VLM-based judge for evaluating anonymization results. 
If no meaningful anonymization is observed, the judge assigns a score of 0; otherwise, it follows the defined rubics to assess the extent to which instance-specific identity attributes have been modified.
}
\label{fig:prompt_judge}
\end{figure*}